\documentclass[11pt]{article}

\usepackage[final]{acl}
\usepackage{subcaption}
\usepackage{amsmath}
\usepackage{times}
\usepackage{latexsym}
\usepackage{cleveref}
\usepackage[T1]{fontenc}

\usepackage[utf8]{inputenc}

\usepackage{microtype}
\usepackage{makecell}
\usepackage{inconsolata}

\newcommand{\cmark}{\textcolor[rgb]{0.0, 0.6, 0.0}{\ding{51}}} 
\newcommand{\xmark}{\textcolor[rgb]{0.7, 0.0, 0.0}{\ding{55}}} 

\usepackage[utf8]{inputenc} 
\usepackage[T1]{fontenc}    
\usepackage{hyperref}       
\usepackage{url}            
\usepackage{booktabs}       
\usepackage{amsfonts}       
\usepackage{nicefrac}       
\usepackage{microtype}      
\usepackage[table]{xcolor}  
\usepackage{graphicx}       
\usepackage{multirow}       
\usepackage{amssymb}
\usepackage{tcolorbox}
\tcbuselibrary{breakable}
\usepackage{listings}

\definecolor{rqBlueBg}{HTML}{EAF4FF}
\definecolor{tzBlueHeader}{RGB}{78,160,205}
\definecolor{tzBlueHeader2}{RGB}{105,185,225}
\definecolor{tzBlueBorder}{RGB}{115,190,225}
\definecolor{tzBlueFill}{RGB}{232,246,252}
\definecolor{rqBlueBorder}{HTML}{6AADE4}
\tcbuselibrary{breakable}
\usepackage{listings}
\usepackage{threeparttable}
\newcommand{\memguardbold}{\textsc{\textbf{MemGuard}}}
\newcommand{\memguard}{\textsc{{MemGuard}}}
\DeclareTextCommand{\textquotedbl}{OT1}{\char`\"}

\lstdefinestyle{jsonTiny}{
  basicstyle=\ttfamily\scriptsize,
  breaklines=true,
  breakindent=0pt,
  columns=fullflexible,
  keepspaces=true,
  showstringspaces=false,
  upquote=true,
  frame=none
}

\tcbuselibrary{skins,breakable}
\usepackage{enumitem}
\newtcolorbox{block}[1][]{%
  enhanced,
  breakable,
  colback=white,
  colframe=black!85,
  boxrule=1.4pt,
  arc=7pt,
  left=4mm,right=4mm,top=4mm,bottom=3mm,
  before skip=10pt, after skip=10pt,
  #1
}

\definecolor{mygreen}{RGB}{60,110,80}
\usepackage{pifont}
\NewDocumentCommand{\ember}
{ mO{} }{\textcolor{mygreen}{\textsuperscript{\textit{Ember}}\textsf{\textbf{\small[#1]}}}}
\NewDocumentCommand{\jy}
{ mO{} }{\textcolor{blue}{\textsuperscript{\textit{Jiayu}}\textsf{\textbf{\small[#1]}}}}
\definecolor{groupgray}{gray}{0.92}

\newenvironment{itemize*}%
 {\leftmargini=20pt\begin{itemize}%
  \setlength{\itemsep}{3pt}%
  \setlength{\parskip}{0pt}%
  }%
 {\end{itemize}} 

\NewDocumentCommand{\yuji}
{ mO{} }{\textcolor{teal}{\textsuperscript{\textit{Yuji}}\textsf{\textbf{\small[#1]}}}}

\NewDocumentCommand{\jeongh}
{ mO{} }{\textcolor{brown}{\textsuperscript{\textit{Jeonghwan}}\textsf{\textbf{\small[#1]}}}}

\NewDocumentCommand{\cheng}
{ mO{} }{\textcolor{orange}{\textsuperscript{\textit{Cheng}}\textsf{\textbf{\small[#1]}}}}

\NewDocumentCommand{\heng}
{ mO{} }{\textcolor{red}{\textsuperscript{\textit{Heng}}\textsf{\textbf{\small[#1]}}}}

\title{\memguard: Preventing Memory Contamination in Long-Term Memory-Augmented Large Language Models}

\author{%
 Hyeonjeong Ha$^{1}$, Jeonghwan Kim$^{1}$, Cheng Qian$^{1}$, \\ \textbf{Jiayu Liu}$^{1}$, \textbf{William M. Campbell}$^{3}$, \textbf{Yue Wu}$^{3}$, \\ \textbf{Yuji Zhang}$^{1}$, \textbf{Kathleen McKeown}$^{2}$, \textbf{Dilek Hakkani-Tür}$^{1}$, \textbf{Heng Ji}$^{1}$\\
$^{1}$University of Illinois Urbana-Champaign, $^{2}$Columbia University, $^{3}$Capital One\\
\texttt{\{hh38, hengji\}@illinois.edu}  \\
}

\begin{document}

\maketitle
\renewcommand{\thefootnote}{\fnsymbol{footnote}}
\footnotetext[2]{Project Page: \url{https://hyeonjeongha.github.io/MemGuard/}.}

\begin{abstract}
Memory-augmented large language models extend reasoning beyond a fixed context window by maintaining long-term memory across interactions. However, existing memory systems often collapse stable user facts, episodic events, and behavioral rules into a shared space, allowing functionally distinct memories to be retrieved and used as interchangeable evidence. We identify this failure mode as \textit{heterogeneous memory contamination}, where context-specific events become overgeneralized claims, or semantically relevant but functionally incompatible memories mislead generation. To this end, we introduce \memguardbold, a type-aware memory framework that preserves functional memory boundaries during memory construction and retrieval. It assigns each memory an explicit functional role at write time, maintains relations across type-isolated memories, and selectively composes evidence only from necessary memory types, reducing contamination from irrelevant or functionally incompatible evidence.
Across hallucination and long-horizon conversation benchmarks, \memguard{} improves memory reliability by up to 28.27\% while retrieving up to 1.23$\times$ fewer memory tokens than prior methods. These results suggest that reliable long-term reasoning depends on principled organization and selective use of heterogeneous memory.

\end{abstract}

\section{Introduction}

Recent memory-augmented large language models (LLMs) move beyond single-context reasoning by persisting knowledge across interactions and reusing it over long time horizons~\citep{wu2024longmemeval, du2024perltqa, ma2024agentboard,park2023generative, shridhar2020alfworld, maharana2024evaluating, liu2025costbench}. This capability is central to personalization and long-horizon reasoning~\citep{zhong2024memorybank, packer2023memgpt, chhikara2025mem0, wang2025mirix, xu2025mem, yan2025memory}, but it also creates a new reliability risk: once noisy, unsupported, or misstructured knowledge is written to memory, it can be repeatedly retrieved and reused. Thus, hallucinations do not arise only at generation time; they can accumulate across the memory cycle, including writing and retrieval.

\begin{figure*}
    \centering
    \includegraphics[width=\linewidth]{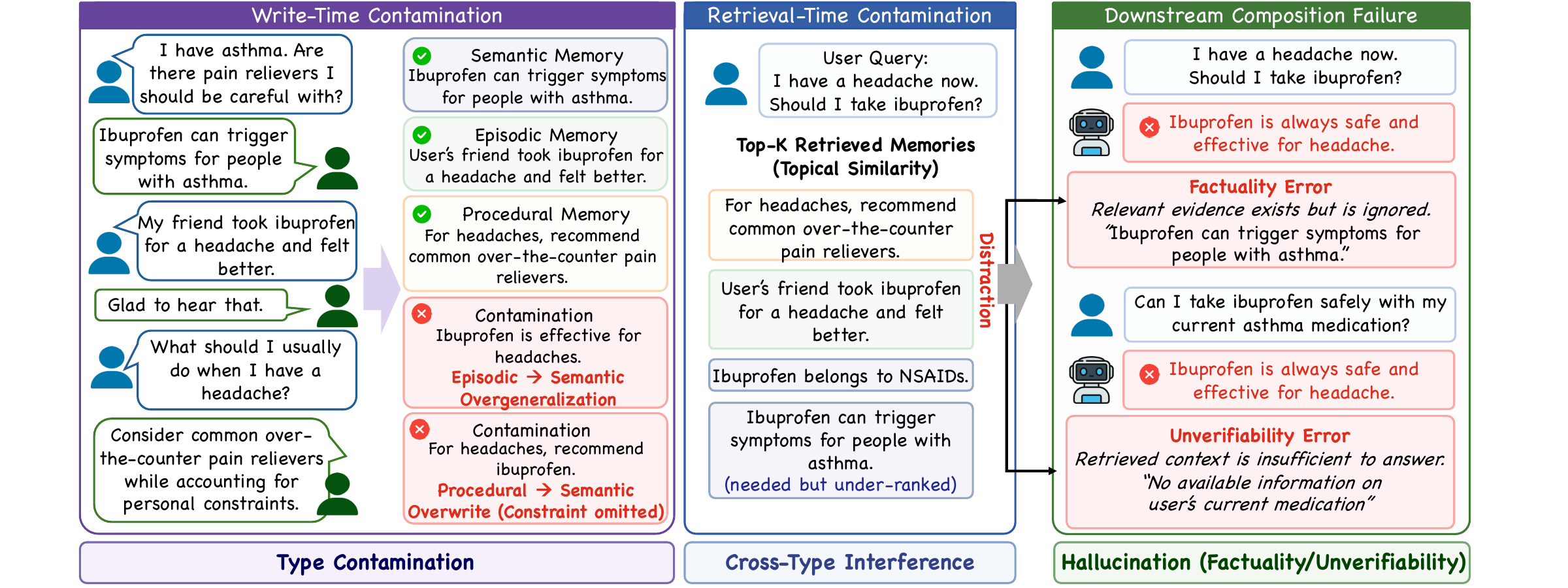}
    \caption{\textbf{Heterogeneous memory contamination.} Weak functional boundaries cause heterogeneous memories, including semantic constraints, episodic observations, and procedural guidance, to be stored, retrieved, and composed as interchangeable evidence. This contamination propagates across the memory writing and retrieval, leading to persistent hallucinations and degraded reasoning quality. 
    }
    \label{fig:hmc}
\end{figure*}
A key source of risk is that conversational memory is functionally heterogeneous: stable facts, event-specific observations, and behavioral rules may be topically related while serving different evidential roles. For example, a memory system may store: (i) a semantic constraint, \textit{[Ibuprofen can trigger symptoms for people with asthma.]}; (ii) an episodic observation, \textit{[User's friend took ibuprofen for headache and felt better.]}; and (iii) a procedural recommendation, \textit{[Given headaches, recommend common over-the-counter pain relievers.]}. Although these memories concern the same topic, they should not be used interchangeably within the same retrieval context. When we disregard the functional distinctions among the evidence, heterogeneous memories can interfere during memory writing and retrieval, a failure mode we refer to as \textbf{heterogeneous memory contamination} (\S\ref{sec:heterogeneous_memory_contamination}). As illustrated in \Cref{fig:hmc}, the contamination can occur when an episodic observation is overgeneralized into an unsupported fact for the user, such as \textit{[Ibuprofen is effective for headaches.]}, while omitting the asthma-related constraint at write-time. At retrieval time, when a query such as \textit{[I have a headache now. Should I take ibuprofen?]} retrieves the episodic success case and procedural recommendation due to topical similarity, but under-ranks the semantic constraint required for safe reasoning. The model may then compose these memories into an unsupported answer~\citep{hong2024so, ha2025mm, jin2025long, park2024toward}. 

Our preliminary analysis on LoCoMo~\citep{maharana2024evaluating} supports this view (\S\ref{sec:heterogeneous_memory_contamination}): \textit{unverifiability errors}, where the model provides responses despite the presence of insufficient evidence, rather than abstaining, are predominantly associated with write-time contamination (97.7\%), while \textit{factuality errors}, where the answer is supported by the conversation history but the model generates an incorrect response, are frequently associated with retrieval-time contamination (63.8\%). However, existing memory-augmented LLMs rarely treat memory-type boundaries as a reliability mechanism. They either treat it as a monolithic memory store and retrieve via semantic similarity~\citep{packer2023memgpt, zhong2024memorybank}, or introduce structural variations to the memory \citep{memobase2025, hu2026evermemos, jiang2026magma, xu2025mem, li2025memos, yu2026agentic, yan2025memory} to stop at optimizing memory utility or retrieval effectiveness. 


To address this gap, we propose \memguardbold, a type-aware memory framework that treats functional boundaries as reliability constraints across memory writing, retrieval, and evidence composition. This design is motivated by cognitive theories of long-term memory, which posit functionally specialized systems that can be selectively recruited and composed according to the reasoning goal~\citep{tulving1972episodic, eustache2008mnesis}. \memguard{} operationalizes this principle for memory-augmented LLMs by separating memories according to their functional roles and controlling how they are later retrieved and composed. At write time, it performs \textit{type-aware memory reorganization}, decomposing conversational content into type-specific atomic memories and recording their dependencies in a relational knowledge graph (\S\ref{sec:write_time}). At retrieval time, it performs \textit{dynamic memory routing}, selecting memory types compatible with the query and composing evidence through the relational graph (\S\ref{sec:write_time}). By disentangling functional distinctions during writing and enforcing type-compatible evidence composition during retrieval, \memguard{} reduces type contamination and cross-type interference, thereby mitigating persistent hallucinations in memory-augmented LLMs.


Experiments on memory hallucination benchmarks and long-horizon conversational tasks show that \memguard{} substantially improves memory reliability. On HaluMem~\cite{chen2025halumem}, \memguard{} achieves 89.53\% (+28.27\%) anti-hallucination accuracy and 71.49\% (+9.38\%) memory update correctness. On LoCoMo~\citep{maharana2024evaluating}, our framework retains competitive performance against state-of-the-art methods, while retrieving \textit{21\% fewer memory tokens}. These results indicate that selectively retrieving the right memory types by preserving functional boundaries is more effective than scaling retrieval indiscriminately. To summarize, our contributions are threefold:

\begin{itemize}[topsep=-1.5pt, leftmargin=10pt, itemsep=-2pt]
    \item We formulate \textit{heterogeneous memory contamination} as a lifecycle-level reliability failure in which weak functional boundaries allow errors to arise and propagate across memory writing, storage, retrieval, and composition.
    \item We propose \memguard{}, a unified framework for contamination control that coordinates type-aware knowledge decomposition , type-isolated memory writing, query-adaptive routing, and relational evidence composition composition to preserve functional boundaries end to end.
    \item We show that \memguard{} improves hallucination robustness while maintaining utility and with fewer retrieved memory tokens.
\end{itemize}

\section{Related Work}

\subsection{Memory-Augmented LLMs}
Memory-augmented LLMs support sustained interactions and long-horizon reasoning, with existing methods broadly falling into four paradigms (\Cref{tab:memory_comparison}). \textit{Flat semantic memory} stores conversational chunks or memories in vector storage and retrieves them via semantic similarity in retrieval-augmented generation (RAG)~\citep{zhong2024memorybank, chhikara2025mem0, memobase2025}. Although efficient, these methods treat history as an unordered set of propositions, making heterogeneous facts prone to inadvertent merging and write-time interference. \textit{Structured and graph-based memory} models the relationships between events, entities, and concepts~\citep{jiang2026magma, rasmussen2025zep, gutierrez2024hipporag, xu2026structmem}, but often retrieves mixed episodic and semantic nodes via similarity or topological search, leaving systems vulnerable to retrieval-time contamination.
\textit{Cognitive-inspired and hierarchical memory} adopts human-like divisions~\citep{packer2023memgpt, kang2025memory, hu2026evermemos, li2025memos, sumers2023cognitive}, such as working, episodic, semantic, and procedural memories, yet commonly relies on static type assignment and mixed retrieval, weakening boundaries between memory types. 
\textit{Agentic and dynamic memory} transforms static storage with context-adaptive, model-driven decisions~\citep{xu2025mem, yan2025memory, yu2026agentic}, but without explicit boundary preservation, it may amplify write-time contamination and uncontrolled associative loops. In contrast, \memguard{} addresses the above-mentioned limitations by preserving the functional boundaries throughout the memory lifecycle. \looseness=-1

\subsection{Reliability in Memory-Augmented LLMs}
Hallucination remains a central challenge in memory-augmented LLMs on long-horizon conversation tasks. Prior work on RAG shows that noisy, conflicting, or weakly relevant retrieved contexts can degrade consistency and induce hallucinations~\citep{park2024toward, hong2024so, ha2025mm, liu2026naacl}. These risks amplify when retrieval operates over accumulated histories and persistent memory stores. HaluMem~\citep{chen2025halumem} shows that hallucinations can emerge across the memory lifecycle, including writing, retrieval, and reasoning, where incorrect knowledge can persist and reinforce erroneous response over time. Others examine retrieval interference and conversational drift from noisy or conflicting memory retrieval~\citep{wu2025sgmem,xu2025mem}. However, existing methods mostly mitigate hallucination after unreliable knowledge has been stored or retrieved, overlooking the \textit{heterogeneity} of knowledge items that differ in memory types.
\section{Heterogeneous Memory Contamination}
\label{sec:heterogeneous_memory_contamination}
\paragraph{Formulation} Conversational memory is inherently heterogeneous, spanning episodic events, semantic facts, and procedural behaviors that serve distinct roles in downstream reasoning. When these functional boundaries are weak, topically related but functionally incompatible memories may be incorrectly updated, retrieved, or composed together. As a result, transient events may be stored as stable facts, or anecdotal evidence can override explicit constraints. We refer to this failure mode as \textbf{heterogeneous memory contamination}: the degradation of memory reliability caused by insufficient functional separation among memory types.

We characterize this contamination across the memory lifecycle (\Cref{fig:hmc}): \textbf{Write-time contamination} occurs when memory construction stores incomplete, outdated, fabricated, or overgeneralized knowledge, causing unsupported content to persist. \textbf{Retrieval-time contamination} occurs when semantically related but functionally unsuitable memories are retrieved together, introducing cross-type noise that obscures the evidence needed for the query. \textit{Composition failures} are typically downstream consequences of write- or retrieval-time contamination: the model incorrectly integrates contaminated or insufficient evidence, allowing irrelevant, conflicting, or weakly grounded memories to distort the final response.

\begin{figure}[t]
    \centering
    \begin{subfigure}[t]{0.49\linewidth}
        \centering
        \includegraphics[width=\linewidth]{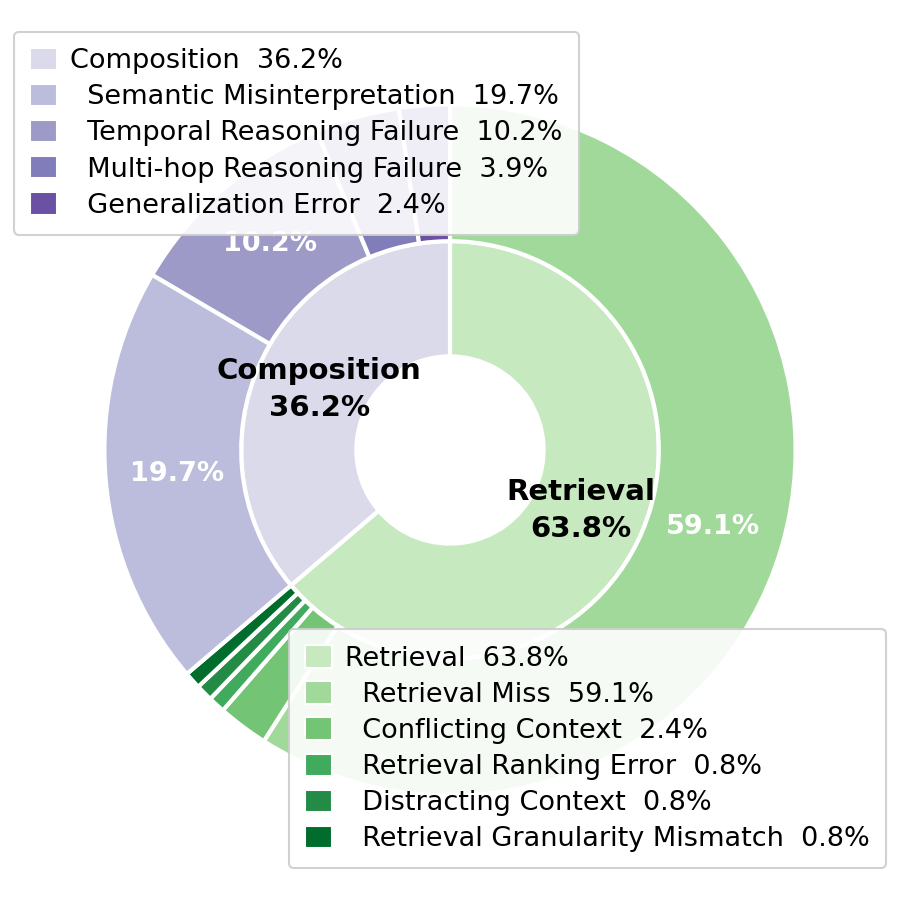}
        \vspace{-5mm}
        \caption{Factuality error.}
        \label{fig:factual_hmc}
    \end{subfigure}
    \hfill
    \begin{subfigure}[t]{0.49\linewidth}
        \centering
        \includegraphics[width=\linewidth]{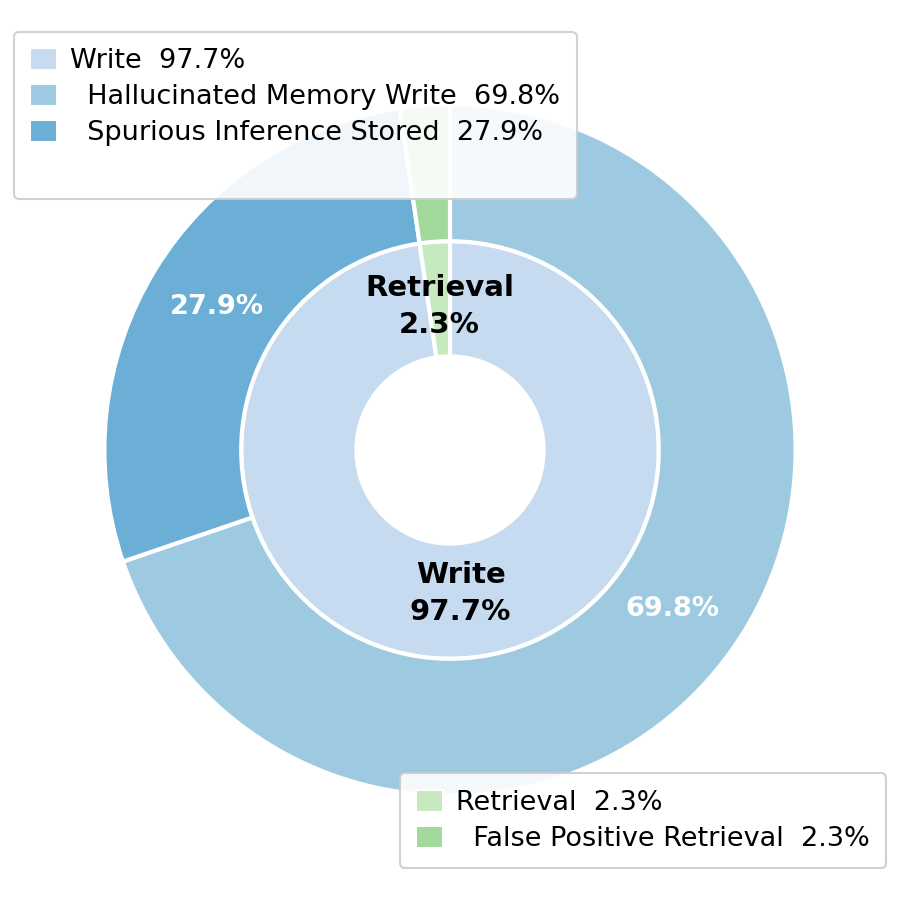}
        \vspace{-5mm}
        \caption{Unverifiability error.}
        \label{fig:hallu_hmc}
    \end{subfigure}
    \vspace{-2mm}
     \caption{{Error analysis across distinct hallucinations.}}
     \vspace{-4mm}
    \label{fig:error_analysis}
\end{figure}

\paragraph{Analysis} We conduct a systematic error analysis on LoCoMo~\citep{maharana2024evaluating}. We categorize failures into (i) \textit{factuality errors}, where the correct answer is recoverable from the conversation but the system generates an incorrect response, and (ii) \textit{unverifiability errors}, where the conversation lacks sufficient evidence, and the model should abstain but instead answers, following prior work~\citep{huang2025survey}. We analyze these errors along two orthogonal dimensions. First, a rule-based decision flow localizes each error to the earliest failing lifecycle stage, which are writing, retrieval, or composition, by comparing the query, ground-truth and generated answers, retrieved context, complete memory storage, and original conversation. For example, when the required evidence exists in storage but is absent from retrieval, the error is assigned to retrieval time. \texttt{GPT-5.2} then assigns the fine-grained subtype within that stage. This rule-guided procedure limits annotation subjectivity, while a human validation study yields 93\% agreement with the resulting fine-grained labels. Second, an independent contamination annotation determines whether the failure involves interference among semantic, episodic, and procedural memories. Full details are provided in Appendix~\ref{appendix:hmc}.

\begin{figure*}
    \centering
    \includegraphics[width=\linewidth]{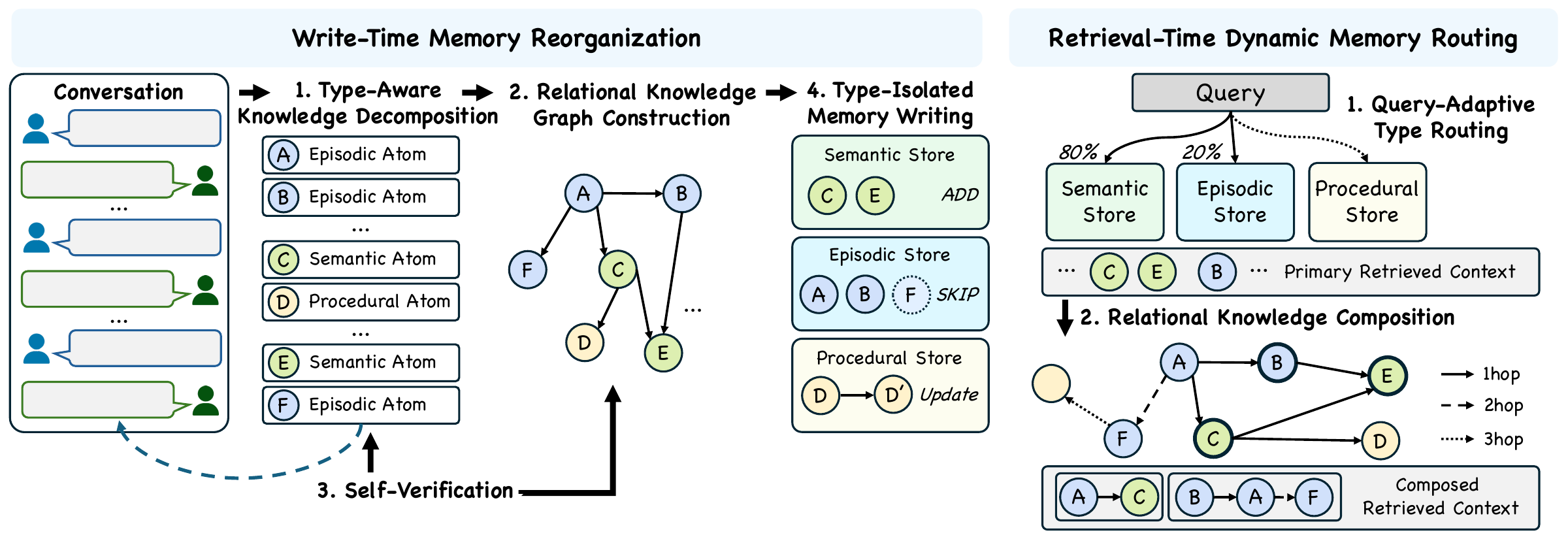}
    \vspace{-0.2in}
    \caption{\textbf{Overview of \memguard{}}. At write time, \memguard{} reorganizes a conversation into atomic knowledge units, constructs directed relations among them, verifies missing information, and writes each atom to a type-isolated memory store. At retrieval time, the model routes queries adaptively to relevant memory types and selectively composes retrieved atoms via a relational knowledge graph, reducing cross-type interference. By preserving functional boundaries, \memguard{} prevents heterogeneous memory contamination.}
    \vspace{-0.1in}
    \label{fig:concept_fig}
\end{figure*}
The lifecycle analysis reveals distinct stage-wise patterns (\Cref{fig:error_analysis}): 97.7\% of unverifiability errors are localized to writing, where unsupported or overgeneralized content enters memory, whereas 63.8\% of factuality errors are localized to retrieval, where available evidence is missed or improperly prioritized. Because stage localization alone does not establish heterogeneous memory contamination, we separately find that 55.6\% of factuality errors involve \textit{cross-type retrieval}, where incompatible memories overshadow correct evidence, and 66.6\% of unverifiability errors involve \textit{cross-type collision}, where weakly grounded memories make an unanswerable query appear answerable. These results identify \textit{cross-type interference} as a substantial source of persistent memory failures and motivate \memguard{} to preserve functional boundaries during writing and routing while restoring necessary cross-type dependencies through controlled relational composition.

\section{\memguard{}}
We introduces a lifecycle-level contamination-control pipeline that coordinates typed atomic decomposition, type-isolated storage, query-adaptive routing, and relational knowledge composition as mutually reinforcing reliability controls. At write time, decomposition establishes functional boundaries and isolated storage prevents cross-type overwriting; at retrieval time, routing restricts search to functionally relevant stores, while relational composition restores necessary cross-type dependencies through explicit relations. Together, these mechanisms preserve functional distinctions end to end, enabling controlled reasoning across heterogeneous memories without treating them as interchangeable evidence.

\subsection{Write-Time Memory Reorganization} 
\label{sec:write_time}
To mitigate heterogeneous memory contamination, \textsc{MemGuard} reorganizes conversational memory before storage. Given a conversation $D$, the write-time procedure converts raw dialogue into type-specific memory atoms, verifies their coverage, links related atoms through a relational graph, and stores them in type-isolated memory stores. This design enforces functional separation at the storage level while preserving explicit dependencies needed for downstream retrieval and reasoning. Formally, \textsc{MemGuard} maintains type-isolated memory stores: $
\mathcal{M}=\{\mathcal{M}_{\tau}\}_{\tau \in \mathcal{T}}$, where $\mathcal{T}=\{\texttt{semantic},\texttt{episodic},\texttt{procedural}\}$, and a relational knowledge graph $\mathcal{G}=(\mathcal{V},\mathcal{E})$, where nodes correspond to memory atoms and edges encode typed dependencies among them.

\paragraph{Type-Aware Knowledge Decomposition}
The LLM decomposes $D$ into a set of non-overlapping memory atoms $A = \operatorname{Decompose}(D)=\{a_j\}_{j=1}^{N}.
$
Each atom captures a single memory unit and is assigned exactly one functional type:
\[
a_j = (\text{title}_j,\text{details}_j,\tau_j,t_j),
\quad
\tau_j \in \mathcal{T},
\]
where $\text{title}_j$ is the short description of $a_j$, $\text{details}_j$ includes actual memory content, and $t_j$ is the absolute timestamp derived from the conversation. The single-type constraint prevents heterogeneous knowledge from being compressed into a shared memory representation. 

\paragraph{Self-Verified Extraction}
Because a single extraction pass may miss relevant information, \textsc{MemGuard} applies a self-verification step before writing. Given the conversation $D$ and the initial atom set $A$, the verifier identifies information in $D$ that is not covered by any atom in $A$. Let $\Delta(D,A)$ denote the recovered missing atoms. The final atom set is $A' = A \cup \Delta(D,A).$ Recovered atoms must satisfy the same constraints as the initial decomposition: each must be atomic, non-overlapping, and tied to a single memory type. We use a single pass to balance coverage improvement with inference cost and to avoid over-extraction or spurious memories that may accumulate under repeated verification.

\paragraph{Relational Knowledge Graph}
Although atoms are isolated by type, downstream reasoning often requires composing related facts, events, and procedures. To preserve such dependencies without merging heterogeneous content, \textsc{MemGuard} constructs a directed typed graph over the atom set:
$\mathcal{V}=\{v_j \mid v_j \equiv a_j,\ a_j \in A'\}$.
Edges encode typed semantic relations:
$\mathcal{E}
=
\{(v_i,v_j,r)\mid v_i,v_j\in \mathcal{V},\ r\in\mathcal{R}\},$ where the relation type $r$ is selected from the relation taxonomy.
A directed edge $(v_i,v_j,r)$ is added when atom $a_i$ relates to atom $a_j$ under relation $r$. For retrieval-time traversal, \textsc{MemGuard} also stores inverse edges:
\[
(v_i,v_j,r)\in\mathcal{E}
\Rightarrow
(v_j,v_i,\texttt{inverse\_}r)\in\mathcal{E}.
\]
Thus, $\mathcal{G}$ preserves cross-atom dependencies, while the atom contents remain type-isolated.

\paragraph{Type-Isolated Memory Writing}
Finally, each atom $a_j \in A'$ is routed to its corresponding typed stores $\mathcal{M}_{\tau_j}$. The atom is compared only with existing top-$N$ (we set $N=20$) relevant memories in the same store (i.e., type-local comparison), and the model assigns one write operation: 
$
o_j \in \{\texttt{ADD}, \texttt{UPDATE}, \texttt{SKIP}\}.
$
\texttt{ADD} inserts a new memory, \texttt{UPDATE} revises an existing memory within the same type-specific store, and \texttt{SKIP} discards redundant or low-value atoms. By restricting deduplication and updates to $\mathcal{M}_{\tau_j}$, \memguard{} prevents functionally distinct memories from overwriting or absorbing one another. Cross-type relations are instead maintained only through $\mathcal{G}$, enabling controlled relational expansion during retrieval without compromising storage-level functional boundaries. We set $K$=20. Details are in Appendix~\ref{sec:memguard_write}.

\subsection{Retrieval-Time Dynamic Memory Routing}
At retrieval time, \textsc{MemGuard} retrieves memories through two stages: query-adaptive type routing and relational composition. The model allocates the primary retrieval budget across type-isolated stores, restricting retrieval to memory types likely to be useful for the query. The retrieved memories are then expanded over the relational knowledge graph constructed at write time, allowing the system to recover relevant cross-memory dependencies without collapsing functional boundaries.

\paragraph{Query-Adaptive Type Routing}
Given a query $q$, \textsc{MemGuard} uses a prompt-based soft router to estimate the relevance of each memory type and output a confidence distribution over memory types:
$\mathbf{w}
=
\rho_{\mathrm{soft}}(q)$, $
w_{\tau}\geq 0$, 
$
\sum_{\tau\in\mathcal{T}} w_{\tau}=1.
$
Each $w_\tau$ denotes the estimated utility of type $\tau$ for answering $q$. Given a retrieval budget $K$, we allocates a type-specific budget $k_{\tau}$ proportional to $\mathbf{w}$: \looseness=-1
$
k_{\tau}
=
\left\lfloor w_{\tau}K \right\rfloor$, $
\sum_{\tau\in\mathcal{T}} k_{\tau}=K.
$
For each type $\tau$, retrieval is performed only over its corresponding store $\mathcal{M}_{\tau}$. Each $m$ is derived from a memory atom
$
a_j=[\mathrm{title}_j,\mathrm{details}_j,\tau_j,t_j].
$
For retrieval, we represent $m$ by concatenating its title and details:
$
\operatorname{text}(m)
=
[\operatorname{title}(m);\operatorname{details}(m)].
$
We rank all entries $m\in\mathcal{M}_{\tau}$ according to the embedding similarity between $q$ and $\operatorname{text}(m)$ and select the $k_{\tau}$ most relevant entries:
\[
P_{\tau}
=
\operatorname{Top}_{k_{\tau}}
\left\{
m\in\mathcal{M}_{\tau}
:
\operatorname{sim}
\left(
q,\operatorname{text}(m)
\right)
\right\}.
\]
Here, $\operatorname{sim}$ denotes the cosine similarity between the L2-normalized embeddings $\phi(q)$ and $\phi(\operatorname{text}(m))$. We use \texttt{text-embedding-3-small} as $\phi(\cdot)$. The primary retrieval set is
$
P
=
\bigcup_{\tau\in\mathcal{T}}P_{\tau},
\qquad
|P|=K.
$
Unlike uniform retrieval over all stores, this routing mechanism makes retrieval capacity query-adaptive while preserving the type-isolation introduced at write time. Details are included in Appendix~\ref{sec:memguard_retrieve}.

\paragraph{Relational Knowledge Composition}
Type routing improves retrieval precision, but primary retrieval may still miss memories that are weakly similar to the query yet essential through relationships. To recover such memories, \textsc{MemGuard} expands each primary result over the relational graph $\mathcal{G}=(\mathcal{V},\mathcal{E})$. For each $p_i \in P$, let $v_i$ be its corresponding graph node. Starting from $v_i$, breadth-first search (BFS) is performed up to $h_{\max}$ hops and collects reachable nodes:
\[
\mathcal{N}_{i}
=
\left\{
(v',r,d)\ \middle|\ 
\begin{aligned}
&v' \in \operatorname{BFS}(v_i,h_{\max}),\\
&r\in\mathcal{R},\ d\leq h_{\max}
\end{aligned}
\right\}.
\]
where $d$ is the hop distance and $r$ is the relation label along the traversal path. Each reachable node is composed into a relation-aware context entry:
\[
c_{i,v'}
=
p_i
\oplus
([\rightarrow r],v'),
\qquad
(v',r,d)\in\mathcal{N}_{i},
\]
where $\oplus$ denotes concatenation with an explicit relation label. The composed entry is scored by query relevance with hop decay:
\[
\operatorname{score}(c_{i,d,v'})
=
\operatorname{sim}(q,c_{i,d,v'})
\cdot
\lambda^{d-1},
\quad
\lambda\in(0,1),
\]
where $\lambda$ is a hop-decay factor. We set $\lambda=0.85$. The final retrieval context is obtained by reranking all graph-expanded entries and selecting the top-$K$:
\[
C
=
\operatorname{Top}_{K}
\left(
\{c_{i,d,v'} \mid p_i \in P,\ (v',r,d)\in\mathcal{N}_i\},
\operatorname{score}
\right).
\]
The answer-generation LLM is conditioned on the query $q$ and the composed context $C$. Overall, query-adaptive routing prevents irrelevant memory types from dominating retrieval, while graph-guided composition restores cross-memory dependencies through explicit typed relations. This provides controlled cross-type reasoning without weakening the functional boundaries enforced during write-time memory reorganization. The overall framework is described in \Cref{fig:concept_fig}.

\section{Experiment}
\subsection{Experimental Setup}
\paragraph{Baselines} To evaluate our method, we compare against memory-augmented LLM baselines spanning diverse memory management paradigms: (1) \textit{flat semantic memory} methods, including Mem0~\citep{chhikara2025mem0} and Memobase~\citep{memobase2025}, (2) \textit{structured/graph memory} methods, including Mem0-Graph~\citep{chhikara2025mem0}, Zep~\citep{rasmussen2025zep}, and Supermemory~\citep{shah2025supermemory}, (3) \textit{cognitive-inspired/hierarchical memory} approaches, including MIRIX~\citep{wang2025mirix} and MemOS~\citep{li2025memos}, and (4) \textit{agentic memory} methods, such as A-Mem~\citep{xu2025mem}. Together, these baselines cover diverse memory structures, representations, and management.

\noindent\textbf{Datasets} We use HaluMem~\citep{chen2025halumem}, which diagnoses hallucinations in memory-augmented LLMs by testing whether models avoid ungrounded memory writes, recognize insufficient evidence, and resist propagating erroneous information during memory writing and generation. This makes it well-suited for measuring contamination across the memory pipeline. We further evaluate long-term memory in realistic conversational settings, i.e., LongMemEval~\citep{wu2024longmemeval}, LoCoMo~\citep{maharana2024evaluating}, and PerLTQA~\citep{du2024perltqa}, all requiring personalized memory retention and retrieval. Together, these benchmarks measure hallucination robustness and long-horizon memory reasoning across diverse settings.

\begin{table*}[t]
    \caption{\textbf{Hallucination evaluation on HaluMem.} Hallucination is evaluated at three stages: memory extraction, update, and answer generation. R/P denote recall/precision; C/H/O denote Correct/Hallucination/Omission rates; and Acc. denotes anti-hallucination accuracy. Higher is better for R, P, Acc., F1, and C; lower is better for H and O. $^\dagger$ indicates results taken from the original paper.}
    \label{tab:halumem_result}
    \centering
    \vspace{-0.1in}
    \resizebox{0.9\textwidth}{!}{
        \begin{tabular}{l *{11}{c}}
            \toprule
            \textbf{Method} &
            \multicolumn{5}{c}{{Extraction}} &
            \multicolumn{3}{c}{{Update}} &
            \multicolumn{3}{c}{{Generation}} \\
            \cmidrule(lr){2-6} \cmidrule(lr){7-9} \cmidrule(lr){10-12}
            & R $\uparrow$ & Weighted R $\uparrow$ & Target P $\uparrow$ & Acc. (\%) $\uparrow$ & F1 $\uparrow$ & C $\uparrow$ & H ($\downarrow$) & O ($\downarrow$) & C $\uparrow$ & H ($\downarrow$) & O ($\downarrow$)\\
            \midrule
            Mem0$^\dagger$                & 42.91 & 65.03 & 86.26 & 60.86 & 57.31 & 25.50 & 0.45 & 74.02 & 53.02 & 19.17 & 27.81 \\
            Mem0-Graph$^\dagger$          & 43.28 & 65.52 & 87.20 & 61.86 & 57.85 & 24.50 & \textbf{0.26} & 75.24 & 54.66 & 19.28 & 26.06 \\
            Zep$^\dagger$         & -- & -- & -- & -- & -- & 47.28 & 0.42 & 52.31 & 55.47 & 21.92 & 22.62 \\
            MemOS$^\dagger$               & 74.07 & 84.81 & 86.25 & 59.55 & 79.70 & 62.11 & 0.42 & 37.48 & \textbf{67.23} & \textbf{15.17} & \textbf{17.59} \\
            Memobase$^\dagger$            & 14.55 & 25.88 & 92.24 & 32.29 & 25.13 & 5.20  & 0.55 & 94.25 & 35.33 & 29.97 & 34.71 \\
            Supermemory$^\dagger$         & 41.53 & 64.76 & 90.32 & 60.83 & 56.90 & 16.37 & 1.15 & 82.47 & 54.07 & 22.24 & 23.69 \\
            \midrule
            \memguard{}   & \textbf{90.47} & \textbf{93.95} & 98.13 & \textbf{89.53} & \textbf{94.15} & 70.79 & 0.35 & \textbf{28.86} & 59.50 & 16.32 & 24.17 \\
            \memguard (2 hop)  & 90.38 & 93.83 & \textbf{98.16} & 89.49 & 94.11 & \textbf{71.49} & 0.51 & 27.99 & 58.26 & 16.61 & 25.12 \\
            \midrule

        \end{tabular}
        \vspace{-0.5in}
    }
\end{table*}

\begin{table*}[t]
    \caption{\textbf{Utility evaluation on general long-horizon conversation benchmarks.} \# Avg. Token refers to the average number of tokens in retrieved memories. Accuracy (\%) is measured through LLM-as-a-Judge. $^\dagger, *$ indicates results taken from the original paper, while “–” indicates results not reported in the original paper.}
    \label{tab:general_result}
    \centering
    \vspace{-0.1in}
    \resizebox{0.9\linewidth}{!}{
        \begin{tabular}{l *{10}{c}}
            \toprule
            {Method} &
            & \multicolumn{7}{c}{LoCoMo} \\
            \cmidrule(lr){2-8}
            & \# Avg. Token & Single Hop & Multi Hop & Open Domain & Temporal & Adversarial & Avg. & PerltQA & LongMemEval \\
            
            \midrule
            \rowcolor{groupgray}
            \multicolumn{10}{c}{\textit{Base LLM: GPT-4o-mini, Judge LLM: GPT-4o-mini}} \\
            Mem0$^\dagger$              & 1172 & 73.33 & 58.75 & 45.83 & 52.34 & -- & 64.57 & -- & 66.40 \\
            Mem0-Graph$^*$          & 3616 & 65.71 & 47.19 & 75.71 & 58.13 & -- & 68.44 & -- & -- \\
            Zep$^\dagger$                 & 2701 & 66.23 & 52.12 & 33.33 & 54.82 & -- & 59.22 & -- & 63.80 \\
            MemOS             & 1589 & 81.09 & 67.49 & 55.90 & 75.18 & -- & {75.80} & -- & {77.80} \\
            Memobase$^\dagger$  & 2102 & 73.12 & 64.65 & 53.12 & 81.20 & -- & 72.01  & -- & 72.40 \\
            Supermemory$^\dagger$ & 500  & 67.30 & 51.12 & 42.67 & 31.77 & -- & 55.34 & -- & 58.40 \\
            MIRIX$^\dagger$ & -    & 68.22 & 54.26 & 46.88 & 68.54 & -- & 64.33 & -- & 43.49 \\
            A-Mem             & 7244 & 60.64 & 38.54 & 62.78 & 23.68 & 68.83 & 56.34 & {77.04} & 47.00 \\ 
            \midrule
            \memguard{} &  1297 & 73.40 & 48.96 & 75.98 & 64.80 & 89.46 & 75.53 & 75.31 & 60.00 \\ 
            \memguard{} (2 hop) & 1250 & 73.76  & 51.04 & 76.58  & 62.93 & 90.13 & 75.78 & 75.64 & 60.00 \\ 
            
            \midrule
            \rowcolor{groupgray}
            \multicolumn{10}{c}{\textit{Base LLM: GPT-4.1-mini, Judge LLM: GPT-4.1}} \\
            A-Mem             & 7244 & 58.87 & 45.83 &  64.80 & 44.24 & 64.35 & 59.62 & {80.62} & 62.25 \\ 
            \midrule
            \memguard{}  & 1605 & 70.92 & 59.38 & 83.12 & 78.01 & 75.11 & 77.29 & 79.44 & {73.50} \\ 
            \memguard{} (2 hop)  & 1540 & 75.89 & 58.33 & 84.54 & 77.57 & 76.46 & {79.10} & 80.07 & 72.25\\ 
            \midrule
        \end{tabular}
    }
\end{table*}

\paragraph{Implementation Details}
For HaluMem, \memguard{} uses \texttt{GPT-4.1-mini} as the base LLM, \texttt{GPT-4.1} as the LLM-as-a-judge due to the computational cost. Baseline results are taken from HaluMem, which uses the stronger \texttt{GPT-4o} as both base LLM and judge; thus, the comparison is conservative for \memguard{} while reducing cost. We exclude A-Mem and MIRIX because they are not reported in HaluMem and require method-specific APIs. To further control for differences in the base LLM, \Cref{app:additional_exp_halumem} reports an additional evaluation of \memguard{} with \texttt{GPT-4o} under the same evaluation settings on a randomly sampled 20\% subset.  

For utility evaluation on LoCoMo, PerLTQA, and LongMemEval, we follow MemOS~\citep{li2025memos}, and use \texttt{GPT-4o-mini} as both base LLM and judge. We also report results under the HaluMem-consistent setting, using \texttt{GPT-4.1-mini} as the base LLM and \texttt{GPT-4.1} as the judge. Baseline numbers are taken from MemOS \textit{(PerLTQA results are unavailable)}, with A-Mem~\citep{xu2025mem} reproduced under our utility setting.

\paragraph{Evaluation Metrics}
Following HaluMem~\citep{chen2025halumem}, we evaluate memory systems across three tasks in the memory lifecycle: memory extraction, memory updating, and memory question answering. For \textbf{memory extraction}, we report recall, weighted recall, target memory precision, memory accuracy (anti-hallucination), and F1, measuring coverage, factuality, and overall extraction quality. For \textbf{memory updating}, we classify each required update as correct, hallucinated, or omitted. For \textbf{memory question answering}, we evaluate the end-to-end reliability after extraction, updating, retrieval, and generation. Each response is judged against the reference answer and key memory points, and categorized as correct, hallucinated, or omitted. We report the correctness, hallucination, and omission rates using LLM-as-a-Judge for memory updating and question answering.

For utility evaluation, we report answer accuracy using an LLM-as-a-judge, following prior work~\citep{li2025memos}. Accuracy on adversarial queries in LoCoMo is not available for existing works. Details are provided in the Appendix~\ref{sec:eval_detail}.

\subsection{Main Results}

\paragraph{Hallucination Evaluation}

\Cref{tab:halumem_result}
shows that \memguard{} reduces hallucination by improving the upstream memory construction and update. In extraction, \memguard{} achieves the highest anti-hallucination accuracy (\texttt{Acc.}=89.53) and F1 score (94.15), indicating that type-aware writing produces cleaner and more complete memory units. In update, it obtains the highest correctness rate (70.79) and the lowest omission rate (28.86), indicating more reliable memory state management. These gains directly support our hypothesis: preserving functional boundaries reduces the risk of heterogeneous knowledge overwriting, obscuring, or conflicting with one another.

Since HaluMem does not directly evaluate retrieval hallucination, we further assess retrieval quality by comparing the retrieved context with gold memory points for each query $q$ (details in Appendix~\ref{sec:eval_prompt_halumem}). \memguard{} achieves 89.87\% and 90.24\% retrieval accuracy with 1-hop and 2-hop retrieval, respectively, comparable to its extraction recall ($R=90.47$). This suggests that type-aware writing and retrieval improve both memory construction and evidence accessibility. The improvement from 1-hop to 2-hop retrieval further suggests that graph-guided composition can recover relationally relevant memories missed by direct semantic similarity. While \memguard{} does not lead on every generation metric, this is consistent with our scope: the method targets contamination before generation, rather than enforcing generation-time grounding. Moreover, \memguard{} uses \texttt{GPT-4.1-mini}, whereas baselines use the stronger \texttt{GPT-4o}; despite this conservative setting, it maintains competitive generation hallucination and omission rates, showing that reducing write- and retrieval-time contamination can improve downstream memory faithfulness. The additional \texttt{GPT-4o} evaluation in \Cref{app:additional_exp_halumem} achieves 74.53 extraction accuracy and 84.76 F1 under the same evaluation settings on a randomly sampled 20\% subset, confirming that \memguard{}'s extraction gains are not specific to its base LLM.

\paragraph{Utility Evaluation} \Cref{tab:general_result} shows the results on long-horizon memory benchmarks. On LoCoMo, \memguard{} achieves 75.53\% average accuracy under the \texttt{GPT-4o-mini} setting, within 0.27\% of MemOS while using fewer retrieved tokens. Under the setting of \texttt{GPT-4.1-mini}/\texttt{GPT-4.1} for base LLM/judge, it reaches to 77.29\%, surpassing A-Mem while using about 4.5$\times$ fewer retrieved tokens. The gain is most pronounced on adversarial queries, which test whether the system can abstain when evidence is insufficient. \memguard{} outperforms A-mem by 20.63\% under the \texttt{GPT-4o-mini} and by 10.76\% under the \texttt{GPT-4.1-mini} setting. These results suggest that preserving functional memory boundaries improves answerability calibration. Type-aware writing reduces unsupported or functionally unsuitable memories from becoming reusable evidence, while query-adaptive routing limits retrieval to memory types relevant to the query, enabling \memguard{} to answer when sufficient evidence exists and abstain more reliably when it does not.

\section{Analysis}
\paragraph{Type-Aware Memory Reorganization Keeps Functional Boundaries.}

\begin{figure}[t]
    \centering
    \begin{subfigure}{0.48\linewidth}
        \centering
        \includegraphics[width=\linewidth]{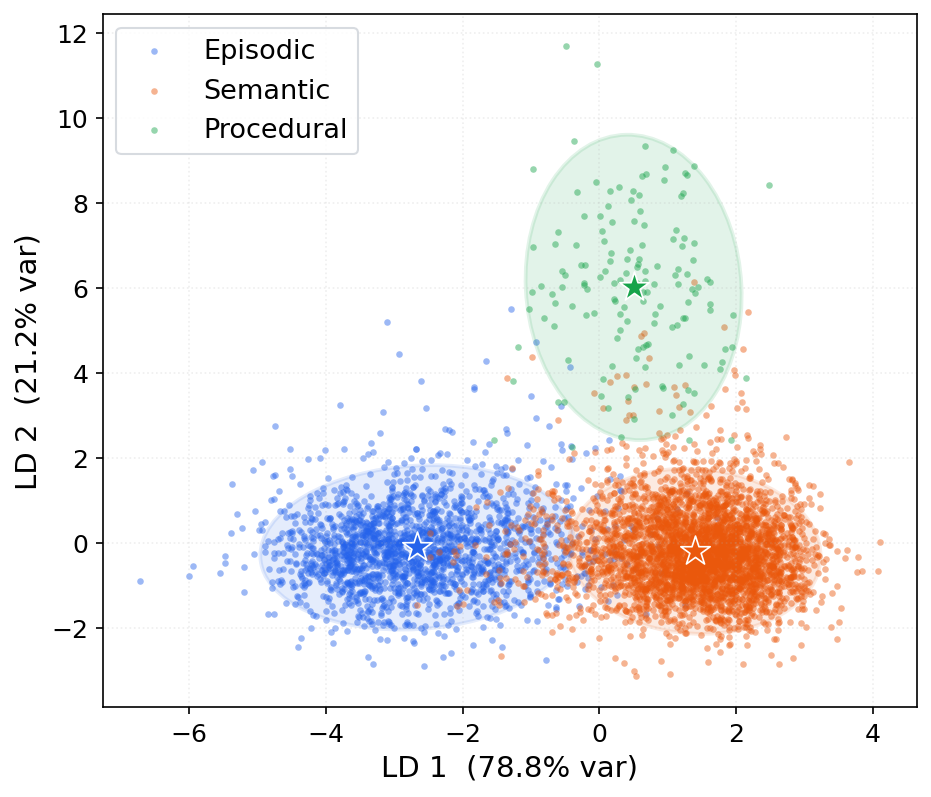}
        \vspace{-0.15in}
        \caption{\small Memory Type Separation.}
        \label{fig:lda_storage}
    \end{subfigure}
    \hfill
    \begin{subfigure}{0.48\linewidth}
        \centering
        \includegraphics[width=\linewidth]{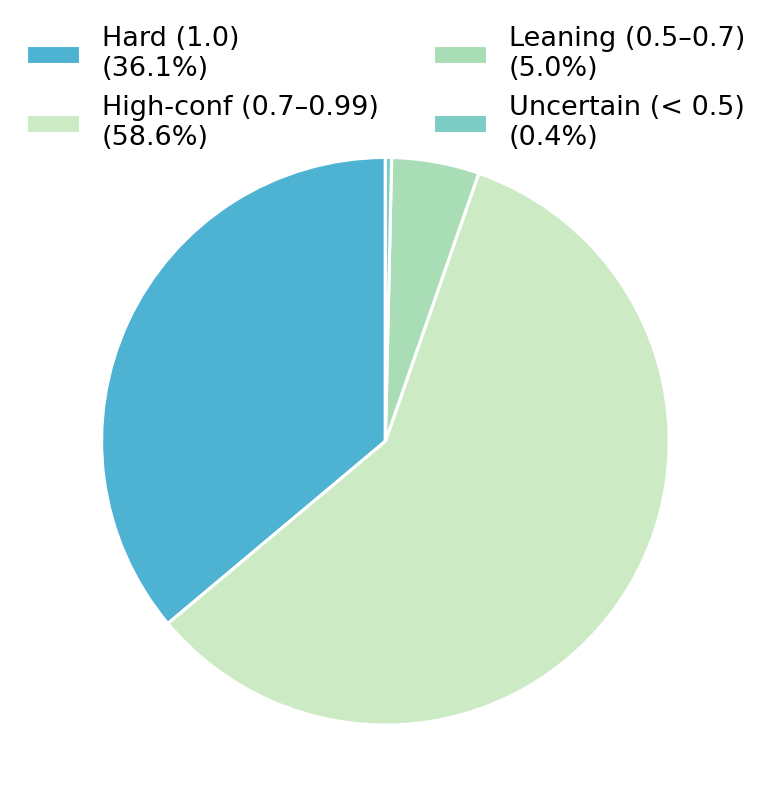}
        \vspace{-0.15in}
        \caption{\small Query Routing Results.}
        \label{fig:routing_confidence}
    \end{subfigure}
    \vspace{-0.1in}
    \caption{\small Analysis of \memguard{} at writing and retrieval.}
    \label{fig:memory_analysis}
\end{figure}

To examine whether \memguard{} mitigates write-time heterogeneous memory contamination, we visualize LoCoMo memories using Linear Discriminant Analysis (LDA) with \texttt{text-embedding-3-small} as an embedding. \Cref{fig:lda_storage} shows that semantic, episodic, and procedural memories form separable clusters, indicating that \memguard{} stores memories in type-consistent regions rather than mixing functionally different knowledge in a shared representation space. The result supports our design: type-aware reorganization preserves functional boundaries. 


\paragraph{Relational Composition Reduces Hallucination Without Sacrificing Utility.}
\begin{figure}
    \centering
    \includegraphics[width=\linewidth]{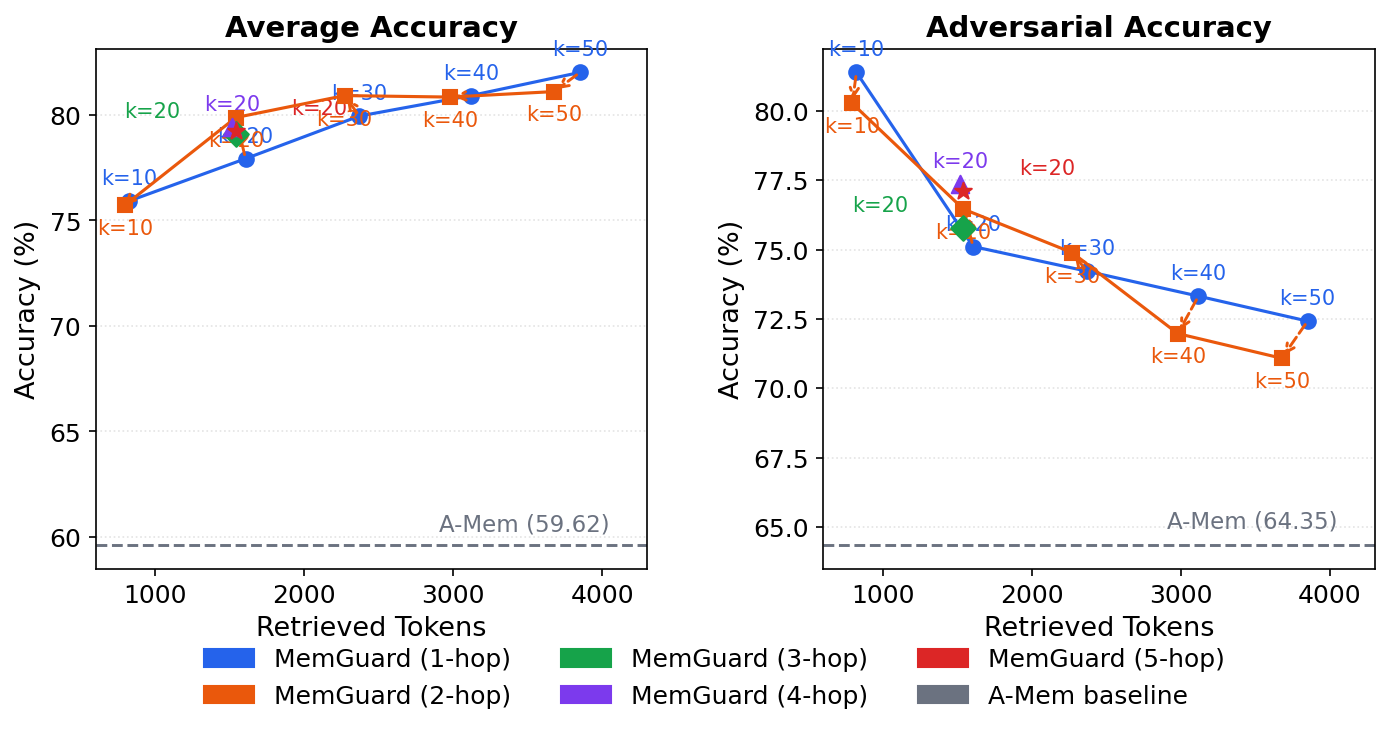}
    \caption{Results with different retrieval budgets and relational hop depth in relational knowledge composition.}
    \vspace{-0.1in}
    \label{fig:hop_analysis}
\end{figure}

We analyze relational knowledge composition on LoCoMo by varying retrieval hop depth and top-$K$ budget. As shown in \Cref{fig:hop_analysis}, increasing $K$ improves average accuracy but lowers adversarial accuracy, indicating that larger contexts recover more answer-supporting evidence while also introducing weakly relevant memories that discourage abstention on ungrounded queries. In contrast, with $K$ fixed at 20, deeper relational expansion maintains utility comparable to the best 2-hop setting while achieving stronger adversarial accuracy. This shows that relational composition improves evidence coverage more selectively than simply increasing retrieval volume, supporting our design of controlled structural expansion for reliable memory retrieval.

\paragraph{Functional Boundaries Require Coordinated Writing, Routing, and Composition.}

            


\begin{table}[t]
    \caption{Results of component ablations on PerLTQA.}
    \vspace{-2mm}
    \label{tab:ablation}
    \centering
    \resizebox{\linewidth}{!}{
        \begin{tabular}{ccccc}
            \toprule
            \makecell{Self-Verified \\ Extraction} &
            \makecell{Type-Isolated \\ Writing} &
            \makecell{Query-Adaptive \\ Routing} &
            \makecell{Relational \\ Composition} &
            \makecell{PerLTQA \\ Accuracy} \\
            \midrule
            \rowcolor{groupgray}
            \cmark & \cmark & \cmark & \cmark & \textbf{79.44} \\
            \cmark & \xmark & \xmark & \xmark & 76.30 \\
            \cmark & \cmark & \xmark & \xmark & 69.71 \\
            \cmark & \cmark & \cmark & \xmark & 74.18 \\
            \xmark & \cmark & \cmark & \cmark & 77.74 \\
            \bottomrule
        \end{tabular}
    }
\end{table}

            


            
\Cref{tab:ablation} reveals a system-level interaction among \memguard{}'s components on PerLTQA. Type-isolated writing alone reduces accuracy from 76.30 to 69.71, indicating that strict isolation can fragment relevant evidence. Query-adaptive routing recovers 4.47\% by selectively retrieving from query-relevant memory types, while relational composition adds a further 5.26\% by restoring necessary cross-type dependencies, yielding the best accuracy of 79.44 and outperforming the shared-store baseline by 3.14\%. Removing self-verified extraction reduces accuracy to 77.74, confirming its role in recovering information missed during initial decomposition. Overall, type-isolated writing establishes functional boundaries, routing exploits these boundaries during retrieval, and relational composition reconnects complementary evidence when needed; their coordinated operation, together with self-verification, produces the strongest performance.

\paragraph{Runtime Efficiency and Reliability Trade-off.}

\begin{table}[t]
    \caption{Comparison of runtime efficiency on LoCoMo. All latency values are reported in milliseconds.}
    \vspace{-2mm}
    \label{tab:runtime_efficiency}
    \centering
    \small
    \setlength{\tabcolsep}{5pt}
    \resizebox{\linewidth}{!}{
        \begin{tabular}{lccccccc}
            \toprule
            Method &
            \makecell{Search\\p99} &
            \makecell{Search\\p90} &
            \makecell{Search\\Mean} &
            \makecell{Search\\Avg.\ Tokens} &
            \makecell{Write\\p99} &
            \makecell{Write\\p90} &
            \makecell{Write\\Mean} \\
            \midrule
            Mem0
                & 4637.0 & 1397.0 & 1089.0 & 1172
                & \textbf{88.0} & 1650.0 & \textbf{88.0} \\
            Memobase
                & 14856.7 & 5974.7 & 2105.2 & 2102
                & 6169.1 & 1533.1 & 1126.2 \\
            Supermemory
                & 2107.6 & 1457.4 & 1057.1 & 500
                & 3467.4 & 1816.9 & 1446.3 \\
            Zep
                & 5348.7 & 614.8 & 571.7 & 2701
                & 375.1 & 254.5 & 239.0 \\
            MemOS
                & \textbf{777.1} & \textbf{528.4} & \textbf{440.5} & 1589
                & 376.4 & \textbf{211.6} & 191.9 \\
            \rowcolor{groupgray}
            \memguard{}
                & 1056.8 & 719.0 & 648.9 & 1297
                & 47716.7 & 27939.2 & 17777.2 \\
            \bottomrule
        \end{tabular}
    }
\end{table}

Following the MemOS~\citep{li2025memos} protocol, we evaluate runtime efficiency on LoCoMo using 10 randomly sampled conversation sessions for writing and 100 randomly sampled queries for search, reporting p99, p90, and mean latency. Write latency covers the complete pipeline, including type-aware decomposition, self-verified extraction, relational graph construction, and type-isolated writing, and is normalized to the time required to add 10 memory entries. Baseline runtime results are taken from \citet{li2025memos}. As shown in \Cref{tab:runtime_efficiency}, \memguard{} achieves a mean search latency of 648.9ms, close to MemOS (440.5ms) and lower than Mem0, Memobase, and Supermemory, indicating that query-adaptive routing and relational composition incur limited query-time overhead. Writing is more expensive, requiring 17.8s on average per 10 entries due to the additional memory-construction stages. This cost yields substantially stronger reliability, including improvements of 27.67\% over the strongest baseline in HaluMem anti-hallucination accuracy and 20.63\% over A-Mem on LoCoMo adversarial queries. Thus, while MemOS remains marginally stronger in average LoCoMo accuracy, \memguard{} offers comparable utility, lower retrieval cost, and stronger contamination resistance, making the additional write-time cost a practical trade-off for reliability-sensitive applications.
\section{Conclusions and Future Work}
We introduced \memguard{}, a framework that improves long-term memory reliability by treating hallucination as a memory-governance failure across writing and retrieval. Our results show that persistent hallucinations often stem from heterogeneous memory contamination, where functionally distinct memories are stored or accessed without sufficient boundary preservation. By enforcing type-aware memory organization and query-adaptive retrieval, it reduces cross-memory interference while enabling structured evidence composition across memory types. Experiments across long-term memory benchmarks show that our method improves memory reliability, strengthens abstention on ungrounded queries, and maintains strong utility with fewer retrieved tokens, suggesting that scalable memory-augmented LLMs should treat memory as a structured and selectively accessed substrate for grounded reasoning. Future work can extend \memguard{} with generation-time grounding to further reduce composition failures.

\clearpage
\section*{Limitation}
\memguard{} improves memory reliability by preserving functional boundaries during writing and retrieval, but it does not directly control generation-time behavior. As a result, although the retrieved memory context is correct and reliable, composition errors may still occur when the base LLM misinterprets or insufficiently grounds its response in the given context. In addition, \memguard{} is implemented as an agentic memory framework built on LLM-based memory construction and retrieval, which may incur additional inference cost compared with fully trained end-to-end memory policies. Finally, our evaluation focuses on long-horizon conversational memory and hallucination benchmarks; future work should validate boundary-preserving memory management in broader settings, such as embodied agents and long-horizon decision-making tasks.

\section*{Acknowledgments}
This research is based upon work supported by the Office of the Director of National Intelligence (ODNI), Intelligence Advanced Research Projects Activity (IARPA), via 560000C260018, CapitalOne-Illinois Center for Generative AI Safety, Knowledge Systems, and Cybersecurity (ASKS), DARPA ITM Program No. FA8650-23-C-7316, DARPA MAGICS Program No. HR0011263E079, Amazon-Illinois Center on AI for Interactive Conversational Experiences (AICE). The views and conclusions contained herein are those of the authors and should not be interpreted as necessarily representing the official policies, either expressed or implied, of IARPA, ODNI, DARPA, or the U.S. Government. The U.S. Government is authorized to reproduce and distribute reprints for governmental purposes notwithstanding any copyright annotation therein. This research used the Delta and DeltaAI advanced computing and data resources, which are supported by the National Science Foundation (award OAC 2320345 and award OAC 2005572) and the State of Illinois. Delta and DeltaAI are joint efforts of the University of Illinois Urbana-Champaign and its National Center for Supercomputing Applications.

\clearpage
{
    \bibliography{reference}
}

\clearpage
\appendix
\begin{table*}[t]
\centering
\small
\vspace{-0.1in}
\caption{Comparison with existing works. Unlike prior methods, \textsc{MemGuard} explicitly preserves functional knowledge boundaries via type-aware memory reorganization at writing and dynamic memory routing at retrieval.
}
\vspace{-0.1in}
\resizebox{\textwidth}{!}{%
\begin{tabular}{lcccccc}
\toprule
\textbf{Method} & \textbf{Memory Structure} & \textbf{Representation} & \textbf{Memory Sharing} & \textbf{Write-time Strategy} & \textbf{Retrieval Strategy} & \textbf{Routing Mechanism} \\
\midrule
\rowcolor{groupgray}\multicolumn{7}{c}{\textit{Flat Retrieval--based Memory}} \\
MemoryBank~\citep{zhong2024memorybank} & Flat / Profile-based & Text summaries + user portrait & Shared & Summarization + memory updating & Similarity-based recall & \xmark \\
Mem0~\citep{chhikara2025mem0} & Flat / Hybrid & Text & Shared & Extract + consolidate salient facts & Similarity search & \xmark \\
MemoBase~\citep{memobase2025} & Profile-based & Structured text profile & Shared & Profile extraction + update & Profile retrieval / similarity search & \xmark \\
\midrule
\rowcolor{groupgray}\multicolumn{7}{c}{\textit{Structured / Graph-based Memory}} \\
Mem0-Graph~\citep{chhikara2025mem0} & Graph-enhanced & Entity-relation graph + text & Shared & Extract + link entities & Graph traversal + similarity search & \xmark \\
MAGMA~\citep{jiang2026magma} & Multi-graph & Semantic/temporal/causal/entity graphs & Shared & Multi-graph construction & Policy-guided graph traversal & Partial \\
Zep~\citep{rasmussen2025zep} & Temporal KG & Text + Entities/Relations & Shared & Incremental extraction \& entity resolution & Graph traversal + similarity & \xmark\\
\midrule
\rowcolor{groupgray}\multicolumn{7}{c}{\textit{Agentic / Dynamic Memory}} \\
A-MEM~\citep{xu2025mem} & Network/Graph-like & Structured notes + links & Shared & Note construction + link generation + evolution & Link-aware retrieval + vector search & \xmark \\
Memory-R1 & Flat & Text entries & Shared & RL-guided operations & RAG + Memory Distillation & \xmark \\
\midrule
\rowcolor{groupgray}\multicolumn{7}{c}{\textit{Cognitive-Inspired / Hierarchical Memory}} \\
MemOS~\citep{li2025memos} & System-level typed memory & Parametric / activation / plaintext memory & Partially shared & Memory governance via MemCube & Managed memory access & Partial \\
EverMemOS~\citep{hu2026evermemos} & Hierarchical / Self-organizing & Text / structured memory units & Shared & Engram-inspired consolidation & Hierarchical / lifecycle-aware retrieval & \xmark \\
MemGPT~\citep{packer2023memgpt} & Hierarchical & Text & Shared & LLM-controlled memory edits & Context paging + archival search & Partial \\
MIRIX~\citep{wang2025mirix} & Typed / Modular & Text memories by type & Partially shared & Static type assignment / consolidation & Mixed type-wise retrieval & \xmark \\
\midrule
\textbf{MemGuard (Ours)} & \textbf{Typed + Dynamic} & \textbf{Atomic units + relational knowledge graph} & \textbf{Isolated} & \textbf{Dynamic routing} & \textbf{Query-aware memory routing} & \cmark \\
\bottomrule
\end{tabular}%
}
\vspace{-0.1in}
\label{tab:memory_comparison}
\end{table*}

\section{Related Work}
\subsection{Memory-Augmented LLMs}
Memory-augmented LLMs have evolved to support sustained, multi-turn interactions and long-horizon reasoning~\citep{wang2023augmenting, hatalis2023memory, wu2024longmemeval, du2024perltqa, ma2024agentboard,park2023generative, shridhar2020alfworld, maharana2024evaluating, liu2025costbench}, with existing methods broadly falling into four paradigms. \textit{Flat semantic memory} stores conversational chunks or memories in vector databases and retrieves them via semantic similarity via retrieval-augmented generation (RAG)~\citep{zhong2024memorybank, chhikara2025mem0, memobase2025}. While being computationally efficient, they treat conversational history as an unordered set of propositions and disparate pieces of knowledge are inadvertently merged, making it prone to interference from merged heterogeneous facts at write-time. \textit{Structured and graph-based memory} explicitly models the relationships between events, entities, and concepts~\citep{jiang2026magma, rasmussen2025zep, gutierrez2024hipporag, xu2026structmem}, yet still retrieves mixed episodic and semantic nodes indiscriminately via similarity or topological search, leaving them susceptible to retrieval-time contamination. 
\textit{Cognitive-inspired and hierarchical memory} replicates the layered storage divisions of the human brain~\citep{packer2023memgpt, kang2025memory, hu2026evermemos, li2025memos}, such as working, episodic, semantic, and procedural memories, but often relies on static type assignment and mixed retrieval, leaving weak boundaries between memory types that lead to cross-type interference.
\textit{Agentic and dynamic memory} transforms static database by employing active, context-adaptive decision processes optimized by the model~\citep{xu2025mem, yan2025memory, yu2026agentic}; however, without explicit boundary preservation, they suffer from exacerbating write-time contamination and raising the risk of uncontrolled associated loops. In contrast, \memguard{} targets the shared limitation across these paradigms: heterogeneous memory contamination. By preserving semantic boundaries throughout the memory lifecycle, \memguard{} aims to reduce cross-type interference and downstream hallucination. Comparison between existing works and \memguard{} is provided in \Cref{tab:memory_comparison}.

\subsection{Relation to MIRIX}
MIRIX~\citep{wang2025mirix} and \memguard{} use typed-memory for different objectives. MIRIX organizes six memory categories to support multimodal personal memory and memory management, whereas \memguard{} targets heterogeneous memory contamination in long-horizon conversations. Accordingly, \memguard{} focuses on episodic, semantic, and procedural memory because they capture the principal functional roles involved in contamination: events, stable facts, and behavioral rules can overwrite, obscure, or conflict with one another when stored without explicit boundaries. The remaining MIRIX categories are orthogonal to this functional taxonomy: Core Memory specifies persistence and visibility, Resource Memory distinguishes external documents and media, and Knowledge Vault imposes privacy and access-control constraints. Content in these categories can still be represented as episodic, semantic, or procedural memory with corresponding persistence, modality, or privacy metadata. Thus, \memguard{}'s three-type taxonomy is not intended to replace MIRIX's architecture, but to provide complementary functional boundaries specifically designed to mitigate cross-type contamination.

\section{Heterogeneous Memory Contamination}
\label{appendix:hmc}
\subsection{Definition of Heterogeneous Memory Contamination}
\begin{table}[h]
\centering
\renewcommand{\arraystretch}{1.25}
\resizebox{\linewidth}{!}{%
\footnotesize
\begin{tabular}{p{2cm} p{0.4cm} p{5cm} p{8cm}}
\toprule
\textbf{Stage} & \textbf{Type} & \textbf{Category} & \textbf{Definition} \\
\midrule

\multirow{5}{*}{Write}
& \multirow{3}{*}{F} & \texttt{memory\_missing}
& Relevant information appears in the conversation but is never written to memory. \\

& & \texttt{abstraction\_error}
& Information is stored with distorted, incomplete, or overly compressed details, losing important specificity. \\

& & \texttt{update\_error}
& A previously stored memory is not updated when new information supersedes or corrects it, resulting in stale knowledge. \\

\cline{2-4}

& \multirow{2}{*}{U} & \texttt{hallucinated\_memory\_write}
& A memory entry is fabricated without grounding in the conversation. \\

& & \texttt{spurious\_inference\_stored}
& An unjustified inference is stored as a fact based on weak or indirect evidence. \\

\midrule

\multirow{7}{*}{Retrieval}
& \multirow{5}{*}{F} & \texttt{retrieval\_miss}
& A correct memory exists in storage but is not retrieved. \\

& & \texttt{retrieval\_ranking\_error}
& The correct memory is retrieved but ranked below less relevant or incorrect memories. \\

& & \texttt{retrieval\_granularity\_mismatch}
& Retrieved memories are too coarse or too fine-grained, preventing correct reasoning. \\

& & \texttt{conflicting\_context}
& Retrieved memories contain contradictory information, and the correct one is not properly selected. \\

& & \texttt{distracting\_context}
& Irrelevant or weakly related memories are retrieved and interfere with reasoning. \\

\cline{2-4}

& \multirow{2}{*}{U} & \texttt{false\_positive\_retrieval}
& Retrieved memories are topically similar but do not contain the required information. \\

& & \texttt{context\_overextension}
& Partial or weak evidence from retrieved memory leads the model to generate unsupported conclusions. \\

\midrule

\multirow{7}{*}{Composition}
& \multirow{4}{*}{F} & \texttt{temporal\_reasoning\_failure}
& The model fails to correctly handle temporal order, recency, or updates across time. \\

& & \texttt{multi\_hop\_reasoning\_failure}
& The model fails to combine multiple retrieved facts to reach a correct conclusion. \\

& & \texttt{semantic\_misinterpretation}
& The model misinterprets the meaning or implication of retrieved content. \\

& & \texttt{generalization\_error}
& The model incorrectly applies or fails to apply knowledge beyond its valid scope. \\

\cline{2-4}

& \multirow{3}{*}{U} & \texttt{parametric\_memory\_intrusion}
& The model relies on pretrained knowledge instead of retrieved memory, overriding relevant context. \\

& & \texttt{unanswerable\_recognition\_failure}
& The model fails to recognize insufficient information and generates an answer instead of abstaining. \\

& & \texttt{hallucinated\_reasoning\_chain}
& The model fabricates a multi-step reasoning process without sufficient grounding in retrieved memory. \\

\bottomrule
\end{tabular}%
}
\caption{Fine-grained taxonomy of memory contamination errors across write-time, retrieval-time, and composition-time, covering two types of hallucinations: factuality errors (F) and unverifiability errors (U).}
\label{tab:error_taxonomy}
\end{table}

\paragraph{Formulation} Memory-augmented LLMs continuously write, retrieve, and compose information from prior interactions to support long-context reasoning and personalization. However, conversational memory naturally contains heterogeneous forms of knowledge, including episodic events, semantic facts, and procedural behaviors, that differ in structure, temporal scope, and intended use. While existing memory systems improve scalability and organization through hierarchical architectures, semantic categories, or adaptive memory management, they still largely rely on shared retrieval spaces where heterogeneous memory types are stored and retrieved together. As a result, semantically distinct knowledge can interfere throughout the memory lifecycle, leading to what we term \textbf{heterogeneous memory contamination}: a failure mode in which weak semantic knowledge boundaries produce noisy memory interactions and degraded reasoning over long interaction horizons.

To better characterize this phenomenon, we organize contamination into three stages of the memory lifecycle: \textit{write-time contamination}, \textit{retrieval-time contamination}, and \textit{composition-time contamination} (\Cref{fig:hmc}). Write-time contamination occurs when heterogeneous knowledge is improperly constructed or updated during memory writing. This includes incomplete abstraction, incorrect updates, unsupported generalization, or fabrication beyond conversational evidence, causing contaminated knowledge to persist in memory over time. Retrieval-time contamination occurs when semantically mismatched memories are retrieved together, introducing noisy or cross-type information that interferes with identifying the relevant evidence for a query. Composition-time contamination occurs when retrieved memories are incorrectly integrated during reasoning, allowing irrelevant episodic details, procedural artifacts, or semantically unrelated memories to distort final answer generation. Although these failures occur at different stages, they frequently propagate throughout the memory pipeline, where early contamination during memory construction degrades downstream retrieval and reasoning. Fine-grained categories of contamination type within each stage are described in \Cref{tab:error_taxonomy}.

\paragraph{Analysis} Based on this formulation, we conduct a systematic error analysis on existing memory frameworks~\cite{chhikara2025mem0} using the LoCoMo~\citep{maharana2024evaluating} benchmark. We divide hallucinations into two categories: \textit{factuality errors}, where the correct answer is recoverable from the conversational history but the system fails to generate it correctly, and \textit{unverifiability errors}, where the conversational history does not contain sufficient evidence for answering and the model should abstain but instead generates unsupported content. Using GPT-5.2 as an LLM-as-a-Judge, we annotate failures across the memory lifecycle and categorize them according to the contamination taxonomy described above, including write-time, retrieval-time, and composition-time failures.

Our annotation results reveal distinct stage-wise failure patterns (\Cref{fig:error_analysis}). Unverifiability errors are predominantly associated with \textit{write-time contamination} (97.7\%), where unsupported, over-generalized, or improperly abstracted knowledge becomes persistently stored in memory and later reinforced across future interactions. In contrast, factuality errors are more strongly associated with \textit{retrieval-time contamination} (63.8\%), where the correct knowledge exists in memory but is not reliably retrieved or appropriately prioritized during reasoning.

To further understand the underlying causes of these failures, we analyze memory interactions across retrieved evidence and identify a common mechanism behind many errors: \textit{cross-type retrieval and collision}. Because heterogeneous memory types are often stored and retrieved within shared memory spaces, semantically distinct knowledge can become entangled during retrieval and reasoning. As a result, topically related but semantically incompatible memories, such as transient episodic observations, procedural instructions, or loosely associated contextual details, are frequently retrieved together and compete during reasoning. We observe that many factuality errors arise not because the correct memory is absent, but because it is overshadowed by irrelevant cross-type memories during retrieval. Similarly, unverifiability errors often emerge when weakly grounded or unsupported memories collide with relevant evidence and become reinforced during generation. Together, these findings suggest that persistent memory failures stem not only from retrieval quality itself, but more fundamentally from weak semantic boundaries between heterogeneous memory types throughout the memory lifecycle.

\subsection{Annotation Pipeline}
To analyze heterogeneous memory contamination, we annotate each incorrect model response with an LLM-based stepwise pipeline. We first distinguish between two failure types: \emph{factuality errors}, where the question has a valid ground-truth answer but the model answers incorrectly, and \emph{hallucinations}, where the question is unanswerable but the model produces a non-abstaining answer. We randomly sample 300 incorrect cases from existing methods for annotation. All annotations are produced by LLM-as-a-judge calls with temperature set to 0, using GPT-5.2 as the LLM. 

\paragraph{Factuality Error Pipeline} 
For factuality errors, we identify the earliest stage at which the correct answer is available. We first check whether the ground-truth answer can be derived from the retrieved context. If so, the failure is attributed either to \emph{retrieval errors}, where the retrieved context is insufficient or misleading despite containing relevant information, or to \emph{composition errors}, where sufficient evidence is retrieved but the model fails to reason over it correctly. If the answer is not recoverable from the retrieved context, we check the full memory storage. When the answer exists in storage but is not retrieved, we label the failure as a \emph{retrieval miss}. If the answer is absent from memory storage, we inspect the original conversation and classify the failure as a write-time error, including \emph{memory missing}, \emph{abstraction error}, or \emph{update error}.

\begin{tcolorbox}[
  enhanced,
  breakable,
  width=0.98\linewidth,
  colback=tzBlueFill,
  colframe=tzBlueBorder,
  boxrule=1.2pt,
  arc=6pt,
  left=5pt,right=5pt,top=4pt,bottom=2pt,
  title={\small Step 1: Is Ground-Truth Answerable from Retrieved Context?},
  coltitle=white,
  colbacktitle=tzBlueHeader2,
  fonttitle=\bfseries,
]
\small
\begin{lstlisting}[style=jsonTiny]
You are an expert evaluator of memory-augmented language models.

## Task
Determine whether the **ground-truth answer** can be correctly derived from the **retrieved context** provided to the model.

## Retrieved Context
{{retrieved_context}}

## Ground-Truth Answer
{{ground_truth}}

## Question
{{question}}

## Instructions
- Answer YES if a careful reader of the retrieved context could produce the ground-truth answer.
- Answer NO if key information needed to arrive at the ground-truth answer is absent or insufficient.
- Do NOT consider whether the model *chose* to use the context correctly - only whether the context *contains* what is needed.

Return JSON only:
```json
{"answerable": true_or_false, "reasoning": "<1-2 sentences citing specific evidence>"}
```
\end{lstlisting}
\end{tcolorbox}

\begin{tcolorbox}[
  enhanced,
  breakable,
  width=0.98\linewidth,
  colback=tzBlueFill,
  colframe=tzBlueBorder,
  boxrule=1.2pt,
  arc=6pt,
  left=5pt,right=5pt,top=4pt,bottom=2pt,
  title={\small Step 1b: Disambiguation - retrieval error vs. composition error},
  coltitle=white,
  colbacktitle=tzBlueHeader2,
  fonttitle=\bfseries,
]
\small
\begin{lstlisting}[style=jsonTiny]
You are an expert evaluator of memory-augmented language models.

The ground-truth answer **can** be derived from the retrieved context, yet the model produced an incorrect answer.
Determine whether the root cause is a **retrieval-quality problem** or a **composition/reasoning problem**.

## Retrieved Context
{{retrieved_context}}

## Model Response
{{model_response}}

## Ground-Truth Answer
{{ground_truth}}

## Question
{{question}}

---

## Distinction

**Retrieval-quality problem** - The composition of *what was retrieved* caused the error:
- A correct memory is present but ranked lower than a misleading one.
- Memories are at the wrong granularity (too coarse or too fine).
- The retrieved set contains two directly contradictory statements.
- An irrelevant memory in the retrieved set distracted the model.
In all these cases, improving the *retriever* - not the *reasoner* - would fix the error.

**Composition/reasoning problem** - The retrieval content was sufficient; the model's *reasoning* failed:
- The model failed to reason about time or sequence correctly.
- The model had all pieces but failed to chain them.
- The model misinterpreted the meaning of a retrieved phrase.
- The model applied a specific fact beyond its intended scope.
In these cases, improving the *reasoner* - not the *retriever* - would fix the error.

Return JSON only:
```json
{"is_retrieval_quality_error": true or false, "reasoning": "<1-2 sentences explaining the deciding signal>"}
```
\end{lstlisting}
\end{tcolorbox}

\begin{tcolorbox}[
  enhanced,
  breakable,
  width=0.98\linewidth,
  colback=tzBlueFill,
  colframe=tzBlueBorder,
  boxrule=1.2pt,
  arc=6pt,
  left=5pt,right=5pt,top=4pt,bottom=2pt,
  title={\small Fine-Grained Retrieval Error Annotation},
  coltitle=white,
  colbacktitle=tzBlueHeader2,
  fonttitle=\bfseries,
]
\small
\begin{lstlisting}[style=jsonTiny]
You are an expert evaluator of memory-augmented language models.

The ground-truth answer **can** be derived from the retrieved context, yet the model's error
stems from a **retrieval-quality problem** - the composition of retrieved memories caused the failure.
Classify the specific retrieval-quality error type.

## Conversation Context
{{conversation}}

## Retrieved Context
{{retrieved_context}}

## Model Response
{{model_response}}

## Ground-Truth Answer
{{ground_truth}}

## Question
{{question}}

---

## ERROR TAXONOMY (retrieval-quality errors)

### retrieval_ranking_error
The correct memory IS present in the retrieved context but is outranked by a less-relevant item
that the model preferentially uses.
- The correct answer is retrievable from the set; model's answer aligns with a weaker, lower-ranked item.
- Example: Retrieved ["User prefers vegetarian meals  (rank 2)", "User ate a burger at a company event (rank 1)"] -> model recommends burger.

### retrieval_granularity_mismatch
Retrieved memories are at the wrong level of detail - too coarse (bundling unrelated facts) or too fine
(splitting linked facts so the inference chain breaks).
- Individual items are not wrong per se, but their granularity prevents the correct inference.
- Example: Chunk A "User takes medication X" + Chunk B "X interacts with alcohol" - only A retrieved -> incomplete picture.

### conflicting_context
The retrieved set contains two or more directly contradictory statements; the model picks the wrong one
or fails to reconcile them.
- Both a correct and an incorrect (or two incompatible) claims appear in the retrieved set.
- Example: Retrieved ["Goal weight is 70 kg", "Target is 80 kg"] -> model uses wrong value.

### distracting_context
The retrieved set contains irrelevant information that is not directly contradictory but misleads the model.
- No direct contradiction; an off-topic or loosely-related fact pulls the answer in the wrong direction.
- Example: "User prefers Python " + "User's team uses Java at work" -> model answers "Java".

---

## Tiebreaker Rules
| Tie | Resolution |
|-----|------------|
| retrieval_ranking_error vs. distracting_context | If the correct answer IS in the set (just lower-ranked) -> ranking_error. If irrelevant noise wins without a correct-answer counterpart -> distracting_context. |
| conflicting_context vs. retrieval_ranking_error | If there is a direct factual contradiction -> conflicting_context. If correct item is simply outweighed by a less-relevant one without direct contradiction -> ranking_error. |

---

## Proposing a Novel Sub-type
If none of the four categories adequately describe the failure:
1. Explain specifically why each existing type fails.
2. Use `snake_case`; describe the mechanism, not the symptom.
3. Set `"is_novel_type": true` and fill `"novel_type_definition"` with one sentence.
4. Set `"confidence"` to `"low"` or `"medium"` - never `"high"` for novel types.

---

Return JSON only:
```json
{
  "error_type": "<retrieval_ranking_error | retrieval_granularity_mismatch | conflicting_context | distracting_context | or new name>",
  "is_novel_type": false,
  "novel_type_definition": null,
  "confidence": "<high | medium | low>",
  "explanation": "<2-4 sentences: what went wrong and why this category fits>",
  "alternative_considered": "<second-most-likely category and why it was ruled out>",
  "evidence": {
    "retrieved_context": "<relevant retrieved snippet>",
    "model_output": "<key incorrect phrase>",
    "ground_truth": "<correct answer>"
  }
}
```
\end{lstlisting}
\end{tcolorbox}

\begin{tcolorbox}[
  enhanced,
  breakable,
  width=0.98\linewidth,
  colback=tzBlueFill,
  colframe=tzBlueBorder,
  boxrule=1.2pt,
  arc=6pt,
  left=5pt,right=5pt,top=4pt,bottom=2pt,
  title={\small Fine-Grained Composition Error Annotation},
  coltitle=white,
  colbacktitle=tzBlueHeader2,
  fonttitle=\bfseries,
]
\small
\begin{lstlisting}[style=jsonTiny]
You are an expert evaluator of memory-augmented language models.

The ground-truth answer **can** be derived from the retrieved context AND the retrieval quality
is adequate. The model's error is a **composition or reasoning failure** - it had what it needed
but failed to produce the correct answer through its reasoning process.
Classify the specific reasoning error.

## Conversation Context
{{conversation}}

## Retrieved Context
{{retrieved_context}}

## Model Response
{{model_response}}

## Ground-Truth Answer
{{ground_truth}}

## Question
{{question}}

---

## ERROR TAXONOMY (composition / reasoning errors only)

### temporal_reasoning_failure
Retrieved context contains time-stamped or sequenced information; model fails to correctly interpret
recency, order, or duration.
- All needed facts are retrieved; error arises from mishandling the time dimension.
- Example: Retrieved ["Lived in NYC in 2021", "Moved to Austin in 2023"] -> model answers "NYC".

### multi_hop_reasoning_failure
All required facts are present in retrieved context; model fails to chain them together to derive
the correct answer.
- No retrieval issue; answer requires combining 2+ facts the model has access to.
- Example: "Alice is Bob's manager" + "Bob works in London" -> "Which office does Alice's report work in?" -> model: "Unknown".

### semantic_misinterpretation
Retrieved context is complete and correctly retrieved; model misunderstands the meaning of a word,
phrase, or implicit reference.
- Error is lexical, pragmatic, or referential - not a retrieval issue.
- Example: "User said they are 'done' with keto" -> model interprets "done" as "finished successfully" rather than "quitting".

### generalization_error
Model correctly retrieves a specific fact but incorrectly generalizes or over-applies it beyond
its intended scope, or fails to apply a general rule to a clear instance.
- Memory and retrieval are correct; error is about scope application.
- Example: "User dislikes cilantro in savory dishes" -> model: "User has no food restrictions."

---

## Tiebreaker Rules
| Tie | Resolution |
|-----|------------|
| multi_hop_reasoning_failure vs. semantic_misinterpretation | Model has all pieces but fails to connect them -> multi_hop. Model misreads the meaning of a single retrieved phrase -> semantic_misinterpretation. |
| generalization_error vs. temporal_reasoning_failure | Scope error is about time/recency -> temporal_reasoning_failure. Scope error is about category/generality, not time -> generalization_error. |

---

## Proposing a Novel Sub-type
If none of the four categories adequately describe the failure:
1. Explain specifically why each existing type fails.
2. Use `snake_case`; describe the mechanism, not the symptom.
3. Set `"is_novel_type": true` and fill `"novel_type_definition"` with one sentence.
4. Set `"confidence"` to `"low"` or `"medium"`.

---

Return JSON only:
```json
{
  "error_type": "<temporal_reasoning_failure | multi_hop_reasoning_failure | semantic_misinterpretation | generalization_error | or new name>",
  "is_novel_type": false,
  "novel_type_definition": null,
  "confidence": "<high | medium | low>",
  "explanation": "<2-4 sentences: what went wrong and why this category fits>",
  "alternative_considered": "<second-most-likely category and why it was ruled out>",
  "evidence": {
    "retrieved_context": "<relevant retrieved snippet>",
    "model_output": "<key incorrect phrase>",
    "ground_truth": "<correct answer>"
  }
}
```
\end{lstlisting}
\end{tcolorbox}

\begin{tcolorbox}[
  enhanced,
  breakable,
  width=0.98\linewidth,
  colback=tzBlueFill,
  colframe=tzBlueBorder,
  boxrule=1.2pt,
  arc=6pt,
  left=5pt,right=5pt,top=4pt,bottom=2pt,
  title={\small Step 2: Is Ground-Truth in Constructed Memory?},
  coltitle=white,
  colbacktitle=tzBlueHeader2,
  fonttitle=\bfseries,
]
\small
\begin{lstlisting}[style=jsonTiny]
You are an expert evaluator of memory-augmented language models.

## Task
Determine whether the **ground-truth answer** can be correctly derived from the **full memory storage**
(all stored memories, not just what was retrieved).

## Full Memory Storage
{{memory_storage}}

## Ground-Truth Answer
{{ground_truth}}

## Question
{{question}}

## Instructions
- Answer YES if any entry in memory storage contains the fact(s) needed to produce the ground-truth answer.
- Answer NO if the needed fact is absent from memory storage entirely.
- Do NOT consider whether retrieval surfaced it - only whether it exists anywhere in storage.

Return JSON only:
```json
{"answerable": true_or_false, "reasoning": "<1-2 sentences citing the specific memory entry or noting absence>"}
```
\end{lstlisting}
\end{tcolorbox}

\begin{tcolorbox}[
  enhanced,
  breakable,
  width=0.98\linewidth,
  colback=tzBlueFill,
  colframe=tzBlueBorder,
  boxrule=1.2pt,
  arc=6pt,
  left=5pt,right=5pt,top=4pt,bottom=2pt,
  title={\small Step 3: Is Ground-Truth in Original Conversation History?},
  coltitle=white,
  colbacktitle=tzBlueHeader2,
  fonttitle=\bfseries,
]
\small
\begin{lstlisting}[style=jsonTiny]
You are an expert evaluator of memory-augmented language models.

## Task
Determine whether the **ground-truth answer** can be correctly derived from the **original conversation**.

## Original Conversation
{{conversation}}

## Ground-Truth Answer
{{ground_truth}}

## Question
{{question}}

## Instructions
- Answer YES if the conversation contains the fact(s) needed to produce the ground-truth answer.
- Answer NO if the conversation does not contain sufficient information.

Return JSON only:
```json
{"answerable": true_or_false, "reasoning": "<1-2 sentences citing the relevant conversation turn or noting absence>"}
```
\end{lstlisting}
\end{tcolorbox}

\begin{tcolorbox}[
  enhanced,
  breakable,
  width=0.98\linewidth,
  colback=tzBlueFill,
  colframe=tzBlueBorder,
  boxrule=1.2pt,
  arc=6pt,
  left=5pt,right=5pt,top=4pt,bottom=2pt,
  title={\small Fine-Grained Write-Time Error Annotation},
  coltitle=white,
  colbacktitle=tzBlueHeader2,
  fonttitle=\bfseries,
]
\small
\begin{lstlisting}[style=jsonTiny]
You are an expert evaluator of memory-augmented language models.

The ground-truth answer **cannot** be derived from either the retrieved context or the full memory
storage, but **can** be derived from the original conversation.
Classify which write-time memory failure caused the information to be lost or distorted.

## Conversation Context
{{conversation}}

## Memory Storage
{{memory_storage}}

## Model Response
{{model_response}}

## Ground-Truth Answer
{{ground_truth}}

## Question
{{question}}

---

## ERROR TAXONOMY (write-time / memory-origin errors)

### memory_missing
Relevant information was present in the conversation but was **never written to memory at all**.
- The correct answer can be derived from the conversation.
- That information is entirely absent from memory storage.
- This is an **omission** failure, not contamination of existing content.
- Example: Conversation "I'm vegetarian and have been for 10 years." Memory: (no dietary info stored) -> memory_missing.

### abstraction_error
Information was stored in memory, but **incorrectly, incompletely, or with distorted meaning** -
without a conflicting update event.
- Memory exists for the topic but key details are wrong, missing, or negated incorrectly.
- There is no temporal conflict - this is a one-time write error.
- Example: Conversation "I'm allergic to shellfish, not just sensitive - I carry an EpiPen."
  Memory: "User is mildly sensitive to shellfish." -> abstraction_error.

### update_error
A new piece of information should have updated an old entry, but the memory system failed -
keeping only the outdated version, only the new version when the old was still valid,
or creating a corrupt merged entry.
- At least **two temporally distinct** conversation turns address the same fact.
- Memory fails to reflect the correct final state.
- Example: Turn 1 "I live in Berlin." Turn 5 "I just relocated to Amsterdam."
  Memory: "User lives in Berlin." -> update_error.

---

## Disambiguation Table
| memory_missing | abstraction_error | update_error |
|----------------|-------------------|--------------|
| Info absent from storage entirely | Info in storage but distorted | Info in storage but stale |
| No memory entry exists for topic | Entry exists, content is wrong | Two competing temporal versions exist |
| Single write never happened | Single write happened incorrectly | Multi-turn update failed |

**Tiebreaker:** Later conversation turn that should overwrite an earlier memory -> update_error.
Single write was simply wrong with no competing version -> abstraction_error.
No write happened at all -> memory_missing.

---

## Proposing a Novel Sub-type
If none of the three categories adequately describe the failure:
1. Explain specifically why each existing type fails.
2. Use `snake_case`; describe the mechanism, not the symptom.
3. Set `"is_novel_type": true` and fill `"novel_type_definition"` with one sentence.
4. Set `"confidence"` to `"low"` or `"medium"`.

---

Return JSON only:
```json
{
  "error_type": "<memory_missing | abstraction_error | update_error | or new name>",
  "is_novel_type": false,
  "novel_type_definition": null,
  "confidence": "<high | medium | low>",
  "explanation": "<2-4 sentences: what went wrong and why this category fits>",
  "alternative_considered": "<second-most-likely category and why it was ruled out>",
  "evidence": {
    "conversation": "<relevant conversation snippet>",
    "memory_storage": "<relevant stored entry or 'N/A - not present'>",
    "model_output": "<key incorrect phrase>",
    "ground_truth": "<correct answer>"
  }
}
```
\end{lstlisting}
\end{tcolorbox}

\paragraph{Unverifiability Error Annotation Pipeline}

For hallucinations, we trace the origin of the fabricated answer. We first check whether the hallucinated content can be linked to memory storage. If so, we label it as a \emph{memory-time contamination}, distinguishing between hallucinated memory writes with no conversational grounding and spurious inferences that over-extend loosely related evidence. If the hallucination is not traceable to storage, we check the retrieved context. Cases grounded in retrieved but non-answering or only partially relevant context are labeled as \emph{retrieval-time contamination}, including false positive retrieval and context overextension. If neither memory storage nor retrieved context explains the hallucination, we attribute the error to \emph{generation-time contamination}, such as parametric memory intrusion, unanswerable recognition failure, or hallucinated reasoning chains.

\begin{tcolorbox}[
  enhanced,
  breakable,
  width=0.98\linewidth,
  colback=tzBlueFill,
  colframe=tzBlueBorder,
  boxrule=1.2pt,
  arc=6pt,
  left=5pt,right=5pt,top=4pt,bottom=2pt,
  title={\small Step 1: Is Hallucination traceable to Constructed Memory?},
  coltitle=white,
  colbacktitle=tzBlueHeader2,
  fonttitle=\bfseries,
]
\small
\begin{lstlisting}[style=jsonTiny]
You are an expert evaluator of memory-augmented language models.

## Task
Determine whether the model's **hallucinated answer** can be traced back to any entry in
**memory storage** - even if that entry is fabricated, inferred, or only loosely related.

This question has **no correct answer** derivable from the original conversation.
The model generated an answer anyway.

## Memory Storage
{{memory_storage}}

## Model's Hallucinated Answer
{{model_response}}

## Question
{{question}}

## Instructions
- Answer YES if any memory entry - even a fabricated, over-inferred, or distorted one -
  could plausibly be the source the model drew on to produce this answer.
- Answer NO if the model's answer has no traceable connection to any entry in memory storage.

Return JSON only:
```json
{"traceable_to_memory": true_or_false, "reasoning": "<1-2 sentences identifying the relevant memory entry or confirming absence>"}
```
\end{lstlisting}
\end{tcolorbox}

\begin{tcolorbox}[
  enhanced,
  breakable,
  width=0.98\linewidth,
  colback=tzBlueFill,
  colframe=tzBlueBorder,
  boxrule=1.2pt,
  arc=6pt,
  left=5pt,right=5pt,top=4pt,bottom=2pt,
  title={\small Fine-Grained Write-Time Annotation},
  coltitle=white,
  colbacktitle=tzBlueHeader2,
  fonttitle=\bfseries,
]
\small
\begin{lstlisting}[style=jsonTiny]
You are an expert evaluator of memory-augmented language models.

## Task
Determine whether the model's **hallucinated answer** can be traced back to anything in the
**retrieved context** - even loosely related or only partially grounding content.

This question has **no correct answer** derivable from the original conversation.

## Retrieved Context
{{retrieved_context}}

## Model's Hallucinated Answer
{{model_response}}

## Question
{{question}}

## Instructions
- Answer YES if any retrieved item - even weak, tangential, or only superficially relevant -
  could plausibly have led the model to generate this answer.
- Answer NO if the model's answer has no traceable connection to the retrieved context.

Return JSON only:
```json
{"traceable_to_retrieval": true_or_false, "reasoning": "<1-2 sentences identifying the retrieved item or confirming no connection>"}
```
\end{lstlisting}
\end{tcolorbox}

\begin{tcolorbox}[
  enhanced,
  breakable,
  width=0.98\linewidth,
  colback=tzBlueFill,
  colframe=tzBlueBorder,
  boxrule=1.2pt,
  arc=6pt,
  left=5pt,right=5pt,top=4pt,bottom=2pt,
  title={\small Step 2: Is Hallucination traceable to Retrieved Context?},
  coltitle=white,
  colbacktitle=tzBlueHeader2,
  fonttitle=\bfseries,
]
\small
\begin{lstlisting}[style=jsonTiny]
You are an expert evaluator of memory-augmented language models.

The model's hallucinated answer can be traced to an entry in **memory storage**.
The question has no correct answer derivable from the original conversation -
the memory entry itself is the source of the hallucination.
Classify which write-time failure introduced the fabricated content.

## Original Conversation
{{conversation}}

## Memory Storage
{{memory_storage}}

## Retrieved Context
{{retrieved_context}}

## Model's Hallucinated Answer
{{model_response}}

## Question
{{question}}

---

## ERROR TAXONOMY (memory-origin hallucination)

### hallucinated_memory_write
The memory system fabricated an entry that has **no grounding** in the original conversation.
- The stored "memory" was invented - never stated, implied, or inferable.
- Distinct from `spurious_inference_stored`: no source statement exists at all.
- Example: Conversation: (no mention of job title). Memory: "User is a senior PM at Google."

### spurious_inference_stored
The memory system stored an **inference or extrapolation** as if it were an explicit fact.
- A loosely related source statement exists, but the stored memory significantly over-extends it.
- The inference chain from source to stored claim is broken or unjustified.
- Distinct from `abstraction_error` (factuality error taxonomy): that type distorts a real, answerable
  fact - here the derived claim was never true to begin with.
- Example: Conversation: "I've been stressed at work lately." Memory: "User has anxiety disorder."

---

**Tiebreaker:** If ANY statement - however distant - could seed the memory -> spurious_inference_stored.
If no such seed exists -> hallucinated_memory_write.

---

Return JSON only:
```json
{
  "error_type": "<hallucinated_memory_write | spurious_inference_stored | or new name>",
  "is_novel_type": false,
  "novel_type_definition": null,
  "confidence": "<high | medium | low>",
  "explanation": "<2-4 sentences>",
  "alternative_considered": "<other category and why ruled out>",
  "evidence": {
    "conversation": "<relevant snippet or 'N/A - no related statement'>",
    "memory_storage": "<the fabricated or over-inferred entry>",
    "model_output": "<key hallucinated phrase>",
    "question": "<the question asked>"
  }
}
```
\end{lstlisting}
\end{tcolorbox}

\begin{tcolorbox}[
  enhanced,
  breakable,
  width=0.98\linewidth,
  colback=tzBlueFill,
  colframe=tzBlueBorder,
  boxrule=1.2pt,
  arc=6pt,
  left=5pt,right=5pt,top=4pt,bottom=2pt,
  title={\small Fine-Grained Retrieval-Time Error Annotation},
  coltitle=white,
  colbacktitle=tzBlueHeader2,
  fonttitle=\bfseries,
]
\small
\begin{lstlisting}[style=jsonTiny]
You are an expert evaluator of memory-augmented language models.

The model's hallucinated answer can be traced to the **retrieved context**, but NOT to memory
storage fabrication. The question has no correct answer derivable from the original conversation.
Classify which retrieval failure caused the model to hallucinate rather than abstain.

## Original Conversation
{{conversation}}

## Memory Storage
{{memory_storage}}

## Retrieved Context
{{retrieved_context}}

## Model's Hallucinated Answer
{{model_response}}

## Question
{{question}}

---

## ERROR TAXONOMY (retrieval-origin hallucination)

### false_positive_retrieval
The retriever surfaced memories topically similar to the question but carrying **no actual answering
information**. The model mistook high retrieval score for factual support.
- Retrieved items are not wrong per se - they are just irrelevant to what the question is asking.
- Example: Q: "What is the user's blood type?" Retrieved: ["User exercises regularly", "User had a health check-up last year"] -> Model: "Blood type is O+."

### context_overextension
The retrieved context contains **weak, partial, or loosely related** evidence; the model fills in
the missing gap with hallucination rather than abstaining.
- Retrieved items have some connection to the question but are insufficient to answer it.
- Distinct from `false_positive_retrieval`: there is a genuine partial connection.
- Example: Q: "What medication does the user take?" Retrieved: ["User mentioned managing a chronic condition"] -> Model: "User takes metformin daily."

---

**Tiebreaker:** Zero factual relationship to the specific claim hallucinated -> false_positive_retrieval.
Partial fact that model over-extended -> context_overextension.

---

Return JSON only:
```json
{
  "error_type": "<false_positive_retrieval | context_overextension | or new name>",
  "is_novel_type": false,
  "novel_type_definition": null,
  "confidence": "<high | medium | low>",
  "explanation": "<2-4 sentences>",
  "alternative_considered": "<other category and why ruled out>",
  "evidence": {
    "retrieved_context": "<the item that misled the model>",
    "model_output": "<key hallucinated phrase>",
    "question": "<the question asked>"
  }
}
```
\end{lstlisting}
\end{tcolorbox}

\begin{tcolorbox}[
  enhanced,
  breakable,
  width=0.98\linewidth,
  colback=tzBlueFill,
  colframe=tzBlueBorder,
  boxrule=1.2pt,
  arc=6pt,
  left=5pt,right=5pt,top=4pt,bottom=2pt,
  title={\small Fine-Grained Composition-Time Error Annotation},
  coltitle=white,
  colbacktitle=tzBlueHeader2,
  fonttitle=\bfseries,
]
\small
\begin{lstlisting}[style=jsonTiny]
You are an expert evaluator of memory-augmented language models.

The model's hallucinated answer cannot be traced to memory storage fabrication or retrieved context.
The hallucination originated entirely at **generation time** (composition-time contamination).
The question has no correct answer derivable from the original conversation.
Classify which generation-level failure caused the model to hallucinate instead of abstaining.

## Original Conversation
{{conversation}}

## Memory Storage
{{memory_storage}}

## Retrieved Context
{{retrieved_context}}

## Model's Hallucinated Answer
{{model_response}}

## Question
{{question}}

---

## ERROR TAXONOMY (generation-time / composition-time contamination)

### parametric_memory_intrusion
The model ignored the retrieved context (empty, irrelevant, or insufficient) and substituted an
answer from its **pretrained parametric knowledge** - treating a personal user question as if it
were a general world-knowledge question.
- The model's pretrained parameters act as an uncontrolled, contaminating memory source: world
  knowledge bleeds into the personal memory system's output.
- Note: This category was previously named `parametric_knowledge_override`; renamed to emphasize
  that parametric parameters constitute a distinct and uncontrolled memory source that contaminates
  personal memory retrieval.
- Example: Q: "What is the user's favorite book?" Retrieved: (nothing relevant) -> Model: "'Atomic Habits'" (popular book from training data).

### unanswerable_recognition_failure
The model failed to detect that the retrieved context is **insufficient** to answer the question
and produced a confident answer instead of abstaining or expressing uncertainty.
- Meta-cognitive calibration failure: the model's "I don't know" trigger did not fire.
- The hallucinated answer may not map to any specific retrieved item or pretrained fact -
  it is confabulated to fill the expected answer slot.
- Example: Q: "How many siblings does the user have?" Retrieved: (empty) -> Model: "The user has two siblings." (stated with confidence).

### hallucinated_reasoning_chain
The model constructs a **multi-step reasoning path** not grounded in retrieved context, fabricating
intermediate steps to bridge from weak/absent evidence to a confident answer.
- Distinct from `multi_hop_reasoning_failure` (factuality error taxonomy): that type fails *with* real
  facts present; here the reasoning chain itself is invented.
- Example: Q: "Does the user prefer remote work?" Retrieved: ["User mentioned commuting takes time"]
  -> Model reasons: "Commuting mentioned -> dislikes commuting -> prefers remote -> Yes, strongly prefers remote."

---

## Tiebreaker Rules
| Tie | Resolution |
|-----|------------|
| parametric_memory_intrusion vs. unanswerable_recognition_failure | Answer resembles well-known general fact -> parametric_memory_intrusion. Confabulated without recognizable factual basis -> unanswerable_recognition_failure. |
| unanswerable_recognition_failure vs. hallucinated_reasoning_chain | No visible reasoning chain -> unanswerable_recognition_failure. Explicit (but fabricated) reasoning steps -> hallucinated_reasoning_chain. |

---

Return JSON only:
```json
{
  "error_type": "<parametric_memory_intrusion | unanswerable_recognition_failure | hallucinated_reasoning_chain | or new name>",
  "is_novel_type": false,
  "novel_type_definition": null,
  "confidence": "<high | medium | low>",
  "explanation": "<2-4 sentences>",
  "alternative_considered": "<second-most-likely category and why ruled out>",
  "evidence": {
    "retrieved_context": "<what was retrieved or '(empty)'>",
    "model_output": "<key hallucinated phrase or reasoning step>",
    "question": "<the question asked>"
  }
}
```
\end{lstlisting}
\end{tcolorbox}

\section{\memguard{} Details}
\subsection{Memory Reorganization at Write-Time}
\label{sec:memguard_write}
\begin{tcolorbox}[
  enhanced,
  breakable,
  width=0.98\linewidth,
  colback=tzBlueFill,
  colframe=tzBlueBorder,
  boxrule=1.2pt,
  arc=6pt,
  left=5pt,right=5pt,top=4pt,bottom=2pt,
  title={\small Atomic Knowledge \& Relation Extraction \& Knowledge Routing},
  coltitle=white,
  colbacktitle=tzBlueHeader2,
  fonttitle=\bfseries,
]
\small
\begin{lstlisting}[style=jsonTiny]
You are an expert knowledge extraction, relationship analysis, and routing system. Your job is to:
1. Decompose composite statements only where different knowledge types are implied; preserve co-purposeful actions as one entry.
2. Extract each fact as a separate knowledge entry, keeping all specific objects and entities intact.
3. Identify relationships between knowledge entries.
4. Route each entry to the correct memory type.

### CONTEXT
Conversation Timestamp: {{conversation_timestamp}}

New Messages:
{{messages}}

---

### PHASE 1: EXTRACT
Scan the conversation for every useful piece of knowledge. Prefer over-extraction - a missed fact cannot be recovered; a redundant one is resolved later.

**What to extract** (capture all that apply):
- Identity & demographics: name, age, occupation, location, education, background
- Personality & traits: self-described style, emotional patterns, decision tendencies, communication style
- Preferences & tastes: food, entertainment, tech tools, travel, aesthetics, communication channels
- Values & beliefs: ethical principles, priorities, worldview, attitudes toward money/work/life
- Goals & plans: short-term tasks, long-term ambitions, learning objectives, financial/health goals
- Constraints & challenges: budget limits, active problems, fears, non-negotiables, frustrations
- Social relationships: family, partner, friends, colleagues, pets - with names and key facts
- Health & wellbeing: conditions, medications, allergies, exercise habits, diet, sleep
- Possessions & resources: devices, vehicles, subscriptions, financial resources, creative tools
- Skills & expertise: professional skills, years of experience, languages, areas actively learning
- Projects & work: active projects (name, goal, status, collaborators), responsibilities, affiliations
- Routines & habits: recurring daily/weekly behaviors, work habits, financial habits, creative habits
- Commitments & obligations: promises to named people, appointments with dates, ongoing duties
- Life events: past/future occurrences, milestones, first-time experiences, reactions
- Opinions & evaluations: ratings, recommendations, positive/negative/mixed reactions to products/people
- Emotional states & reactions: expressed feelings tied to specific events or situations; emotional tone toward named people or places; significant emotional turning points volunteered by the speaker
- Current life situation: life phase, recent major transitions, environmental/seasonal context
- Domain knowledge: custom vocabulary, frameworks, named systems or tools they personally built

**Decomposition - split composite statements before extracting:**
Scan every statement for facts that belong to different memory types bundled together. Split only when types diverge; keep co-purposeful actions together.

| Split trigger | How to split |
|---|---|
| Past event + derived stable belief or trait | Episodic entry for the event; semantic entry for the belief |
| Occurrence + timeless motivational/emotional explanation | Separate entry for the occurrence; separate entry for the stable motivation |
| Stable preference + the one-time event that revealed it | Episodic for the event; semantic for the preference |
| Two or more subjects sharing one predicate | One entry per subject |

**Do NOT split when:**
- Multiple actions share the same subject, context, and type (e.g., "will do A and B for C" -> one entry)
- Multiple reasons/effects all belong to the same type -> merge into one entry

Example - *"Melanie painted a lake sunrise last year and finds painting a fun way to express herself, get creative, and relax."*
-> atom 0: Melanie painted a lake sunrise (time-anchored occurrence)
-> atom 1: Melanie uses painting to express herself, get creative, and relax (stable multi-purpose belief - kept together since all purposes are semantic)

Counter-example - *"I'm going to launch a website and run ads for my limited edition hoodie line."*
-> ONE episodic entry (same type, same subject, same context - do not split)

**What to skip:**
- Common public knowledge (e.g. "Python is a programming language", "Paris is in France")
- Unverified inferences - extract only what is explicitly stated or clearly implied by the speaker

**Completeness rules:**
- Never omit named objects, products, places, quantities, or people from an entry
- Preserve exact names, numbers, dates, units, and quoted phrases
- Never bundle a time-anchored event and the stable belief it implies in one entry
- Never mix information about different people in the same entry
- If uncertain or ambiguous, set "uncertain": true

**Time normalization:**
- `time` records when the event actually occurred (YYYY-MM-DD, YYYY-MM, or YYYY) - NOT the conversation timestamp
- Resolve relative expressions ("next month", "last year") to absolute dates using the Conversation Timestamp
- In `details`, include both the resolved event time and the conversation timestamp so the entry is fully self-contained.
  - Format: `<resolved_time>: <fact> (mentioned on <conversation_timestamp>)`
  - Example: conversation timestamp 2022-03-27, speaker says "I've been playing drums for a month":
      -> resolved event start: 2022-02
      -> `time`: "2022-02"
      -> `details`: "John has been playing drums for a month (mentioned on 2022-03-27); he described it as tough but fun."
- If the day-of-week is needed but not known (e.g. "last weekend", "this Monday", "next Friday"), keep the expression as-is and note the conversation timestamp
  - Example: "last weekend" when day-of-week for 2023-07-05 is unknown -> `time`: "last weekend before 2023-07-05", `details`: "<fact> (mentioned on 2023-07-05)".
- If no time is mentioned, use the Conversation Timestamp for `time`

---

### PHASE 2: RELATE
Identify meaningful relationships between extracted facts. Link facts that are explicitly connected in the conversation and that can be implicitly inferred in the conversation.

Use exactly these relation labels:

| Label        | Meaning                                                             |
|---|---|
| `supports`   | source supports, evidences, or is required by / informed by target  |
| `instance_of`| source is a concrete occurrence of the target concept               |
| `derived_from` | source stable fact was concluded from or triggered by target experience |
| `leads_to`   | source event directly caused or led to target event                 |
| `context_for`| source provides background or context for target                    |
| `elaborates` | source adds detail or specificity to target                         |
| `contradicts`| source conflicts with or updates target                             |

Rules:
- Each link is directional: source -> target follows the relation's meaning
- One atom may appear in multiple links
- If no relationships exist, output "links": []

---

### PHASE 3: ROUTE
Assign exactly one type to every extracted atom:

- **semantic**: timeless and stable facts about a person or a world
- **episodic**: time-anchored past/future events or experiences 
- **procedural**: recurring behaviors or routines 

**Guardrails:**
- Past, long-time ago experience is anchored to time -> episodic, NOT semantic
- A one-time description of how to do something -> episodic, NOT procedural
- Procedural is ONLY for behaviors the person performs on a recurring basis

---

### OUTPUT FORMAT
Return a single JSON object:

{
  "atoms": [
    {
      "id": 0,
      "type": "semantic" | "episodic" | "procedural",
      "title": "Short label (<=10 words)",
      "details": "Full details including resolved event time and conversation timestamp where applicable",
      "time": "YYYY-MM-DD|YYYY-MM|YYYY",
      "uncertain": true|false
    }
  ],
  "links": [
    {
      "source": 0,
      "target": 1,
      "relation": "<relation_label>",
      "reasoning": "<one sentence grounded in the conversation>"
    }
  ]
}

### CRITICAL RULES:
1. Return ONLY the JSON object; no preamble, explanation, or trailing text
2. Every extracted fact must appear as exactly one atom
3. Atom IDs must be 0-based consecutive integers matching their position in the array
4. Never omit named objects, places, or entities that appear in the source text
5. "title" must be unique and self-explanatory without surrounding context

\end{lstlisting}
\end{tcolorbox}

\begin{tcolorbox}[
  enhanced,
  breakable,
  width=0.98\linewidth,
  colback=tzBlueFill,
  colframe=tzBlueBorder,
  boxrule=1.2pt,
  arc=6pt,
  left=5pt,right=5pt,top=4pt,bottom=2pt,
  title={\small Memory Operation Assignment},
  coltitle=white,
  colbacktitle=tzBlueHeader2,
  fonttitle=\bfseries,
]
\small
\begin{lstlisting}[style=jsonTiny]
You are a memory operation assignment system. Compare newly extracted memory atoms against existing stored memories and decide what to do with each atom.

### CONTEXT
Conversation Timestamp: {{conversation_timestamp}}

Existing Semantic Memories (compare ONLY for semantic atoms):
{{existing_semantic_memories}}

Existing Episodic Memories (compare ONLY for episodic atoms):
{{existing_episodic_memories}}

Existing Procedural Memories (compare ONLY for procedural atoms):
{{existing_procedural_memories}}

New Memory Atoms:
{{atoms_json}}

---

### PHASE 4: ASSIGN OPERATIONS
For every atom, compare it against existing memories OF THE SAME TYPE and assign exactly one action:

- **ADD**: No existing memory covers this fact. Create a new entry.
- **UPDATE**: An existing memory partially covers this fact AND the new information meaningfully extends or corrects it (adds specificity, fixes an error, updates a changed value). Include `old_memory_id`.
- **SKIP**: The fact is already fully captured by an existing memory. Include `existing_id` (the ID of the overlapping memory). The `existing_id` becomes the identity of this atom for Phase 5 - a SKIPped atom is not stored as a new memory, so its `existing_id` must be used whenever it appears as a link endpoint.

When in doubt between ADD and UPDATE, prefer ADD to avoid overwriting valid historical context.

Compare:
- semantic atoms -> against existing_semantic_memories
- episodic atoms -> against existing_episodic_memories
- procedural atoms -> against existing_procedural_memories

---

### PHASE 5: EXISTING LINKS
Links between new atoms were already captured in Phase 2 (RELATE). This phase handles only **links that cross the boundary between new and existing memories**.

Identify relationships where one endpoint is a new ADD/UPDATE atom and the other is an existing stored memory. This covers:
- A new ADD/UPDATE atom that elaborates, contradicts, or provides context for an existing memory
- A SKIPped atom whose `existing_id` is meaningfully connected to another ADD/UPDATE atom - use the SKIP's `existing_id` as the endpoint, not the atom index

**Do NOT create links where both endpoints are new atoms** - those belong to Phase 2 and are already recorded.

Use exactly these relation labels:

| Label        | Meaning                                                             |
|---|---|
| `supports`   | source supports, evidences, or is required by / informed by target  |
| `instance_of`| source is a concrete occurrence of the target concept               |
| `derived_from` | source stable fact was concluded from or triggered by target experience |
| `leads_to`   | source event directly caused or led to target event                 |
| `context_for`| source provides background or context for target                    |
| `elaborates` | source adds detail or specificity to target                         |
| `contradicts`| source conflicts with or updates target                             |

Each link needs exactly one source and one target. Choose the correct field:
- ADD/UPDATE atom as endpoint: use `source_atom` or `target_atom` (integer atom id)
- Existing memory as endpoint (including a SKIP's `existing_id`): use `source_existing_id` or `target_existing_id` (memory id string)

**Do NOT use `source_atom`/`target_atom` for SKIPped atoms** - they are not stored as new memories. Always reference them via `source_existing_id`/`target_existing_id`.

Only create links explicitly grounded in the conversation. If none exist, output "existing_links": [].

---

### OUTPUT FORMAT
Return a single JSON object:

{
  "operations": [
    {"atom_id": 0, "action": "ADD"},
    {"atom_id": 1, "action": "UPDATE", "old_memory_id": "<existing_memory_id>"},
    {"atom_id": 2, "action": "SKIP", "existing_id": "<existing_memory_id>"}
  ],
  "existing_links": [
    {
      "source_atom": 0,
      "target_existing_id": "<existing_memory_id>",
      "relation": "<relation_label>",
      "reasoning": "<one sentence grounded in the conversation>"
    },
    {
      "source_existing_id": "<existing_memory_id>",
      "target_atom": 1,
      "relation": "<relation_label>",
      "reasoning": "<one sentence>"
    },
    {
      "source_existing_id": "<skip_atom_existing_id>",
      "target_atom": 2,
      "relation": "<relation_label>",
      "reasoning": "<atom 2 is new ADD/UPDATE; the SKIP atom is referenced via its existing_id>"
    }
  ]
}

### CRITICAL RULES:
1. Return ONLY the JSON object - no preamble, explanation, or trailing text
2. Every atom must appear in exactly one operation
3. SKIP operations must include `existing_id`
4. UPDATE operations must include `old_memory_id`

\end{lstlisting}
\end{tcolorbox}

\begin{tcolorbox}[
  enhanced,
  breakable,
  width=0.98\linewidth,
  colback=tzBlueFill,
  colframe=tzBlueBorder,
  boxrule=1.2pt,
  arc=6pt,
  left=5pt,right=5pt,top=4pt,bottom=2pt,
  title={\small Self-Check Memory Extraction},
  coltitle=white,
  colbacktitle=tzBlueHeader2,
  fonttitle=\bfseries,
]
\small
\begin{lstlisting}[style=jsonTiny]
You are a memory extraction auditor. A first-pass extraction has already been run on the conversation below. Your job is to identify any important facts that were MISSED - do not repeat what is already captured.

### CONTEXT
Conversation Timestamp: {{conversation_timestamp}}

New Messages:
{{messages}}

---

### Already Extracted Atoms (do NOT duplicate these):
{{existing_atoms_json}}

---

### TASK
Carefully re-read the conversation and list any important facts NOT yet captured above.

**Extraction Rules:**
- Only report genuinely new facts - skip anything already covered by an existing atom
- Skip common public knowledge and unverified inferences
- Preserve exact names, numbers, dates, and entities
- Split composite facts that span different types (episodic event vs. semantic belief)
- Use `time` for when the event occurred (YYYY-MM-DD, YYYY-MM, or YYYY); resolve relative expressions using the Conversation Timestamp
- New atom IDs must start at {{next_id}} and increment consecutively

**Relation Rules:**
- Links may connect new additional atoms to each other OR to existing atoms (using their existing IDs)
- Use exactly these relation labels:
  | Label        | Meaning                                                             |
  |---|---|
  | `supports`   | source supports, evidences, or is required by / informed by target  |
  | `instance_of`| source is a concrete occurrence of the target concept               |
  | `derived_from` | source stable fact was concluded from or triggered by target experience |
  | `leads_to`   | source event directly caused or led to target event                 |
  | `context_for`| source provides background or context for target                    |
  | `elaborates` | source adds detail or specificity to target                         |
  | `contradicts`| source conflicts with or updates target                             |

**Routing Rules:**
Assign exactly one type to every extracted atom:

- **semantic**: timeless and stable facts about a person or a world
- **episodic**: time-anchored past/future events or experiences 
- **procedural**: recurring behaviors or routines 

If nothing was missed, return `{"additional_atoms": [], "additional_links": []}`.

---

### OUTPUT FORMAT
{
  "additional_atoms": [
    {
      "id": {{next_id}},
      "type": "semantic" | "episodic" | "procedural",
      "title": "Short label (<=10 words)",
      "details": "Full details including resolved event time and conversation timestamp",
      "time": "YYYY-MM-DD|YYYY-MM|YYYY",
      "uncertain": true|false
    }
  ],
  "additional_links": [
    {
      "source": <atom_id>,
      "target": <atom_id>,
      "relation": "<relation_label>",
      "reasoning": "<one sentence grounded in the conversation>"
    }
  ]
}

### CRITICAL RULES:
1. Return ONLY the JSON object - no preamble, explanation, or trailing text
2. New atom IDs must start at {{next_id}} - never reuse existing atom IDs
3. Never omit named objects, places, or entities from atom details

\end{lstlisting}
\end{tcolorbox}

\subsection{Dynamic Memory Routing at Retrieval-Time}
\label{sec:memguard_retrieve}

\begin{tcolorbox}[
  enhanced,
  breakable,
  width=0.98\linewidth,
  colback=tzBlueFill,
  colframe=tzBlueBorder,
  boxrule=1.2pt,
  arc=6pt,
  left=5pt,right=5pt,top=4pt,bottom=2pt,
  title={\small Dynamic Routing at Memory Retrieval},
  coltitle=white,
  colbacktitle=tzBlueHeader2,
  fonttitle=\bfseries,
]
\small
\begin{lstlisting}[style=jsonTiny]
You are a memory routing assistant. Given a user query, assign a confidence weight (0.0-1.0) to each memory type that may contain the answer. Weights must sum to 1.0.

Memory types:
- semantic: timeless and stable facts about a person or a world
- episodic: time-anchored past/future events or experiences 
- procedural: recurring behaviors or routines 

Query: {{user_query}}

Return ONLY a JSON object:
{"weights": {"semantic": <float>, "episodic": <float>, "procedural": <float>}}
\end{lstlisting}
\end{tcolorbox}

\begin{tcolorbox}[
  enhanced,
  breakable,
  width=0.98\linewidth,
  colback=tzBlueFill,
  colframe=tzBlueBorder,
  boxrule=1.2pt,
  arc=6pt,
  left=5pt,right=5pt,top=4pt,bottom=2pt,
  title={\small Answer Generation},
  coltitle=white,
  colbacktitle=tzBlueHeader2,
  fonttitle=\bfseries,
]
\small
\begin{lstlisting}[style=jsonTiny]
Retrieved memories: {retrieved context}

Question: {question}

Instructions:
1. Carefully analyze the retrieved memories to find relevant information
2. Consider synonyms and related concepts (e.g., "support group", "activist group" may refer to similar things)
3. If memories mention specific dates/times, use those to answer time-related questions
4. If memories contain contradictory information, prioritize the most recent memory
5. Focus on the content of the memories, not just exact word matches

**For factual questions (What/When/Where/Who):**
- Answer based on direct information in the memories
- If the specific fact is not mentioned, respond: "Not answerable"

**For inference/reasoning questions (Would/Could/Likely):**
- You CAN make reasonable inferences based on related information in the memories

**When to say "Not answerable":**
- If the question asks about a specific person but the memories are about a DIFFERENT person
- If the question asks about an event/action that is NOT mentioned in ANY memories
- If you find information about a similar but DIFFERENT event

Provide a concise, direct answer based on the available information, or state "Not answerable" if the specific information is not present.
\end{lstlisting}
\end{tcolorbox}
 
\section{Evaluation Details}
\label{sec:eval_detail}
\subsection{Dataset}
We use HaluMem~\citep{chen2025halumem}, a benchmark specifically designed to diagnose hallucinations in memory-augmented LLMs. HaluMem systematically probes the model's ability to (i) avoid producing ungrounded memory entries, (ii) correctly identify when sufficient evidence is lacking, and (iii) resist propagating erroneous or hallucinated information during both memory writing and generation. This makes it particularly suitable for measuring contamination across the memory pipeline. To further evaluate long-term memory capabilities in realistic conversational settings, we additionally consider three widely used dialogue benchmarks: (i) LongMemEval~\citep{wu2024longmemeval} that consists of user-assistant chat histories with 400 questions for test split, (ii) LoCoMo~\citep{maharana2024evaluating} that contains 10 human-human conversations between fictional personas grounded in temporal event graphs, including 600 dialogues and 26,000 tokens on average with averagely 200 questions for each conversation, and (iii) PerLTQA~\citep{du2024perltqa} features 141 characters with rich personal profiles, social relationships, and life events that includes 8,593 questions over 30 characters, all requiring personalized memory retention and retrieval.

\subsection{Implementation Details}
For hallucination evaluation on HaluMem~\citep{chen2025halumem}, we follow the original protocol, while evaluating on HaluMem-Medium due to computational cost. Unless otherwise noted, \memguard{} uses \texttt{GPT-4.1-mini} as the base LLM, \texttt{GPT-4.1} as the LLM-as-a-judge, and \texttt{text-embedding-3-small} for the embedding model in retrieval. We retrieve the top-10 memories for memory updating and the top-20 memories for question answering, matching the HaluMem setup. Baseline results are taken from the original HaluMem paper, where \texttt{GPT-4o} serves as both the base LLM and judge. Due to the high cost of \texttt{GPT-4o}, \memguard{} uses \texttt{GPT-4.1-mini} as the base LLM, and this comparison is conservative for \memguard{} while substantially reducing computational cost as \texttt{GPT-4o} is stronger than \texttt{GPT-4.1-mini}. A-Mem and MIRIX are excluded from HaluMem comparisons because they were not reported in the original benchmark and require method-specific APIs.

For utility evaluation on LoCoMo~\citep{maharana2024evaluating} and LongMemEval~\citep{wu2024longmemeval}, we follow prior work~\citep{li2025memos}, using \texttt{GPT-4o-mini} as both the base LLM and the LLM-as-a-judge. By default, \memguard{} retrieves the top-20 memories for memory updating and top-$k$ memories for answer generation, with $k=20$. We further analyze the effect of varying $k$ and the number of retrieval hops. To isolate the effect of model choice, we also report utility results under the HaluMem-consistent setting, using \texttt{GPT-4.1-mini} as the base LLM and \texttt{GPT-4.1} as the judge.

Unless explicitly reproduced, baseline results are taken from their original papers: hallucination results from HaluMem~\citep{chen2025halumem} and utility results from MemOS~\citep{li2025memos}. This is unavoidable because most baselines depend on method-specific APIs beyond the OpenAI API. The only exception is A-Mem~\citep{xu2025mem}, which we reproduce under our utility evaluation setting.

\subsection{Evaluation Metrics}

Following HaluMem~\citep{chen2025halumem}, we evaluate hallucination behavior at three stages: memory extraction, memory updating, and memory question answering. For memory extraction, we assess both coverage and factuality. Let $\mathcal{G}$ denote the set of gold memory points and $\mathcal{M}$ denote the memories extracted by the system. We measure memory completeness using recall, measuring the proportion of reference memories that are successfully extracted by the memory system:
\[
\mathrm{R}
=
\frac{|\mathcal{G}_{\mathrm{matched}}|}{|\mathcal{G}|}.
\]

We also report weighted recall, which accounts for the relative importance of each reference memory, assigning higher weight to more important memories and giving partial credit when a memory is only partially extracted:

\[
\mathrm{Weighted\ R}
=
\frac{\sum_{g_i \in \mathcal{G}} w_i s_i}
{\sum_{g_i \in \mathcal{G}} w_i},
\]

where $w_i$ is the importance weight of gold memory $g_i$, and $s_i \in \{1, 0.5, 0\}$ is the extraction score indicating whether the memory is fully extracted, partially extracted, or omitted. To measure factuality, target precision evaluates the correctness of extracted memories that correspond to gold targets:

\[
\mathrm{Target\ P}
=
\frac{
|\mathcal{M}_{\mathrm{correct}} \cap \mathcal{M}_{\mathrm{target}}|
}{
|\mathcal{M}_{\mathrm{target}}|
}.
\]

Memory accuracy evaluates the correctness of all extracted memories, which assesses whether the extracted memories are factual and free from hallucination:

\[
\mathrm{Acc}
=
\frac{|\mathcal{M}_{\mathrm{correct}}|}
{|\mathcal{M}|}.
\]

We compute memory extraction F1 as the harmonic mean of recall and target precision, which shows the overall performance of the memory extraction task by jointly considering completeness and correctness:

\[
\mathrm{F1}
=
\frac{
2 \cdot \mathrm{R} \cdot \mathrm{Target\ P}
}{
\mathrm{R} + \mathrm{Target\ P}
}.
\]

For memory updating, we evaluate whether the system can correctly modify, merge, or replace existing memories during new dialogues so that consistency is maintained without introducing hallucinations. Following HaluMem, each update is categorized as correct, hallucinated, or omitted. Given $N_{\mathrm{upd}}$ target updates, we report the correctness, hallucination, and omission rates:

\[
\mathrm{C}_{\mathrm{upd}}
=
\frac{N_{\mathrm{correct}}}{N_{\mathrm{upd}}},
\]
\[
\mathrm{H}_{\mathrm{upd}}
=
\frac{N_{\mathrm{hallucinated}}}{N_{\mathrm{upd}}},
\]
\[
\mathrm{O}_{\mathrm{upd}}
=
\frac{N_{\mathrm{omitted}}}{N_{\mathrm{upd}}}.
\]
The \emph{correctness rate} measures the proportion of required updates that are correctly applied. The \emph{hallucination rate} measures the proportion of updates that introduce incorrect or fabricated information. The \emph{omission rate} measures the proportion of required updates that are not applied or are missed by the memory system.
 
For memory question answering, we evaluate the end-to-end reliability of the memory system after memory extraction, updating, retrieval, and answer generation. The system retrieves relevant memories and generates an answer for each question, which is then compared against the reference answer. The \emph{correctness rate} measures the proportion of questions answered correctly. The \emph{hallucination rate} measures the proportion of answers that contain unsupported or incorrect information. The \emph{omission rate} measures the proportion of answers that leave the question unanswered due to missing memories. Given $N_{\mathrm{qa}}$ questions, we compute:

\[
\mathrm{C}_{\mathrm{qa}}
=
\frac{N_{\mathrm{correct}}}{N_{\mathrm{qa}}},
\]
\[
\mathrm{H}_{\mathrm{qa}}
=
\frac{N_{\mathrm{hallucinated}}}{N_{\mathrm{qa}}},
\]
\[
\mathrm{O}_{\mathrm{qa}}
=
\frac{N_{\mathrm{omitted}}}{N_{\mathrm{qa}}}.
\]
Higher values are better for recall, weighted recall, target precision, memory accuracy, F1, and correctness rate, whereas lower values are better for hallucination and omission rates.

For utility evaluation on LoCoMo~\citep{maharana2024evaluating} and LongMemEval~\citep{wu2024longmemeval}, we report answer accuracy using an LLM-as-a-judge, following prior work~\citep{li2025memos}.

\subsection{Hallucination Evaluation Prompts}
\label{sec:eval_prompt_halumem}
We use the prompts provided by HaluMem~\citep{chen2025halumem} for answer generation, memory integrity evaluation, memory update evaluation, and qa generation evaluation.
\begin{tcolorbox}[
  enhanced,
  breakable,
  width=0.98\linewidth,
  colback=tzBlueFill,
  colframe=tzBlueBorder,
  boxrule=1.2pt,
  arc=6pt,
  left=5pt,right=5pt,top=4pt,bottom=2pt,
  title={\small Answer Generation},
  coltitle=white,
  colbacktitle=tzBlueHeader2,
  fonttitle=\bfseries,
]
\small
\begin{lstlisting}[style=jsonTiny]
You are an intelligent memory assistant tasked with retrieving accurate information from conversation memories.

   # CONTEXT:
   You have access to memories retrieved from a conversation history. Each memory may include
   a timestamp, contextual summary, and tags that can help you identify relevant information.

   # INSTRUCTIONS:
   1. Carefully analyze all provided memories
   2. Pay special attention to the timestamps to determine the answer
   3. If the question asks about a specific event or fact, look for direct evidence in the memories
   4. If memories contain contradictory information, prioritize the most recent memory
   5. If there is a question about time references (like "last year", "two months ago", etc.),
      calculate the actual date based on the memory timestamp. For example, if a memory from
      4 May 2022 mentions "went to India last year," then the trip occurred in 2021.
   6. Always convert relative time references to specific dates, months, or years.
   7. Focus only on the content of the memories. Do not confuse character names mentioned
      in memories with the actual users who created those memories.
   8. The answer should be less than 5-6 words.

   # APPROACH (Think step by step):
   1. First, examine all memories that contain information related to the question
   2. Examine the timestamps and content of these memories carefully
   3. Look for explicit mentions of dates, times, locations, or events that answer the question
   4. If the answer requires calculation (e.g., converting relative time references), show your work
   5. Formulate a precise, concise answer based solely on the evidence in the memories
   6. Double-check that your answer directly addresses the question asked
   7. Ensure your final answer is specific and avoids vague time references

   {context}

   Question: {question}

   Answer:
\end{lstlisting}
\end{tcolorbox}

\begin{tcolorbox}[
  enhanced,
  breakable,
  width=0.98\linewidth,
  colback=tzBlueFill,
  colframe=tzBlueBorder,
  boxrule=1.2pt,
  arc=6pt,
  left=5pt,right=5pt,top=4pt,bottom=2pt,
  title={\small Memory Integrity Evaluation},
  coltitle=white,
  colbacktitle=tzBlueHeader2,
  fonttitle=\bfseries,
]
\small
\begin{lstlisting}[style=jsonTiny]
You are a strict **"Memory Integrity" evaluator**.
Your core task is to assess whether an AI memory system has **missed any key memory points** after processing a conversation. This evaluation measures the system's **memory integrity**, i.e., its ability to resist **amnesia** or **omission**.

# Evaluation Context & Data:

1. **Extracted Memories:**
   These are all the memory items actually extracted by the memory system.
   {memories}

2. **Expected Memory Point:**
   The key memory point that *should* have been extracted.
   {expected_memory_point}

# Evaluation Instructions:

1. For each **Expected Memory Point**, search within the **Extracted Memories** list for corresponding or related information. Ignore unrelated items.
2. Based on the following scoring rubric, rate how well the memory system captured the **Expected Memory Point** and provide a detailed explanation.

# Scoring Rubric:

* **2:** Fully covered or implied.
  One or more items in "Extracted Memories" fully cover or logically imply all information in the "Expected Memory Point."

* **1:** Partially covered or mentioned.
  Some information in "Extracted Memories" mentions part of the "Expected Memory Point," but key information is missing, inaccurate, or slightly incorrect.

* **0:** Not mentioned or incorrect.
  "Extracted Memories" contains no mention of the "Expected Memory Point," or the corresponding information is entirely wrong.

# Scoring Notes:

* For **compound Expected Memory Points** (with multiple elements such as person/event/time/location/preference, etc.):

  * All elements correct -> **2 points**
  * Some elements correct / uncertain -> **1 point**
  * Key elements missing or wrong -> **0 points**

* Semantic matching is acceptable; exact wording is **not** required.

* If "Extracted Memories" contains **conflicting information**, assign the **best possible coverage score** and mention the conflict in your reasoning.

* Extra or stylistically different memories do **not** reduce the score; only the coverage of the **Expected Memory Point** matters.

* For uncertain wording ("might," "probably," "tends to," etc.):

  * If the Expected Memory Point is a definite statement, usually assign **1 point**.

* If critical fields (e.g., time, entity name, relationship) are partly wrong but others match -> **1 point**.

  * If all key fields are wrong or missing -> **0 points**.

# Output Format:

Please output your result in the following JSON format:

```json
{{
  "reasoning": "Provide a concise justification for the score",
  "score": "2|1|0"
}}
```
\end{lstlisting}
\end{tcolorbox}

\begin{tcolorbox}[
  enhanced,
  breakable,
  width=0.98\linewidth,
  colback=tzBlueFill,
  colframe=tzBlueBorder,
  boxrule=1.2pt,
  arc=6pt,
  left=5pt,right=5pt,top=4pt,bottom=2pt,
  title={\small Memory Accuracy Evaluation},
  coltitle=white,
  colbacktitle=tzBlueHeader2,
  fonttitle=\bfseries,
]
\small
\begin{lstlisting}[style=jsonTiny]
You are a **Dialogue Memory Accuracy Evaluator.** Your task is to evaluate the **accuracy** of a memory extracted by an AI memory system, based on three given inputs: the dialogue content, the *target (gold)* memory points (the correct annotated memories), and the *candidate* memory to be evaluated. The goal is to output a **structured evaluation result**.

# Input Content

* **Dialogue:**
  {dialogue}

* **Golden Memories (Target Memory Points):**
  The correct memory points pre-annotated for this dialogue in the evaluation dataset.
  {golden_memories}

* **Candidate Memory:**
  The memory extracted by the system to be evaluated.
  {candidate_memory}

# Evaluation Principles and Definitions

### 1) Support / Entailment

* An **information point** (atomic fact) in the candidate memory is considered *supported* if it can be directly stated or semantically entailed (via synonym, paraphrase, or equivalent expression) by the *Dialogue* or *Golden Memories*.
* Only the given dialogue and golden memories can be used for judgment - **no external knowledge** or assumptions are allowed.
  Any information not appearing in or inferable from these two sources is considered *unsupported*.
* Pay careful attention to **negation**, **quantities**, **time**, and **subjects**.
  If the candidate statement contradicts the dialogue or golden memories, it is considered a **conflict**.

### 2) Memory Accuracy Score (integer: 0 / 1 / 2)

* **2 points:** Every information point in the candidate memory is supported by the dialogue or golden memories, with **no contradictions or hallucinations**.
* **1 point:** The candidate memory is *partially correct* (at least one supported information point) but also includes *unsupported* or *contradictory* content.
* **0 points:** The candidate memory is **entirely unsupported or contradictory** to the sources (i.e., a "hallucinated memory").

> Note:
>
> * If a candidate memory contains multiple information points, **any unsupported or contradictory element** prevents a full score (2).
> * If both supported and unsupported/conflicting content appear, assign a score of **1**.

### 3) Inclusion in Golden Memories (Boolean field-level judgment)

**Definition:**

* **Atomic information point:** the smallest factual unit in the candidate memory (e.g., *name = Li Si*, *age = 25*, *location = Beijing*, *preference = coffee*, *budget <= 2000*, *meeting_time = Wednesday 10:00*, *tool = Zoom*, etc.).
* **Field / Slot:** the semantic dimension of an information point (e.g., *name*, *age*, *residence*, *food preference*, *budget*, *meeting time*, *meeting tool*, etc.).

**Judgment Rules (independent of correctness):**

* **true:**
  Every atomic information point in the candidate memory has a corresponding **field** in the golden memories (allowing for synonyms, paraphrases, or equivalent expressions; ignore value, polarity, or quantity differences).

  * Note: A single field in the gold list may match multiple candidate points (e.g., multiple "drink preference" facts can be covered by one "drink preference" field in gold).
* **false:**
  If **any** atomic information point's field in the candidate memory cannot be found in the golden memories, mark as *false*.

**Important Notes:**

* Field matching is restricted to fields that are **explicitly present or semantically recognizable** in the golden memories - no external knowledge may be used to expand the field set.
* Differences in **values** (e.g., "Zhang San" vs. "Li Si"), **polarity** (like/dislike), or **exact number/time** do **not** affect this Boolean judgment.

# Evaluation Procedure

For each candidate memory:

1. **Decompose** it into atomic information points (e.g., name, number, location, preference).
2. For each information point, **search** the dialogue and golden memories for supporting or contradictory evidence.
3. Assign the **accuracy_score** (0 / 1 / 2) according to the rules above.
4. Determine **is_included_in_golden_memories (true/false)**:

   * Identify each information point's field;
   * If *all* fields exist in the golden memories, mark as *true*; otherwise, *false*.
5. Provide a **concise Chinese explanation** in `"reason"`, citing key evidence (short excerpts allowed), and clearly state any unsupported or contradictory parts if applicable.

# Output Format (strictly required)

Output **only one JSON object**, with the following three fields:

* `"accuracy_score"`: `"0"` or `"1"` or `"2"`
* `"is_included_in_golden_memories"`: `"true"` or `"false"`
* `"reason"`: `"brief explanation in Chinese"`

Do **not** include any other text, explanation, or fields.
Do **not** include the candidate memory text inside the JSON.

Please output **only** the following JSON (in a code block):

```json
{{
  "accuracy_score": "2 | 1 | 0",
  "is_included_in_golden_memories": "true | false",
  "reason": "Brief explanation in Chinese"
}}
```
\end{lstlisting}
\end{tcolorbox}

\begin{tcolorbox}[
  enhanced,
  breakable,
  width=0.98\linewidth,
  colback=tzBlueFill,
  colframe=tzBlueBorder,
  boxrule=1.2pt,
  arc=6pt,
  left=5pt,right=5pt,top=4pt,bottom=2pt,
  title={\small Memory Update Evaluation},
  coltitle=white,
  colbacktitle=tzBlueHeader2,
  fonttitle=\bfseries,
]
\small
\begin{lstlisting}[style=jsonTiny]
Your task is to **evaluate the update accuracy** of an AI memory system.
Based on the information provided below, determine whether the system-generated **"Generated Memories"** correctly **includes** the **Target Memory for Update**.

# Background Information

The following information is provided for evaluation:

1. **Generated Memories:**
   This is the list of memory points generated by the system after the current dialogue.
   {memories}

2. **Target Memory for Update:**
   This is the correct, updated version of the memory point that should have been produced - the one we focus on in this evaluation.
   {updated_memory}

3. **Original Memory Content:**
   This is the original version of the target memory before the update.
   {original_memory}

# Evaluation Criteria

Please make your judgment **strictly based on the content update of the "Target Memory for Update."**
Use the following categories:

### Correct Update

* **Generated Memories** **contains all information points** from the "Target Memory for Update," accurately and completely reflecting the intended update.
* **Key fields** (e.g., date, time, values, proper nouns, etc.) must match exactly.
* The **original memory** is effectively replaced or marked as outdated.
* Synonymous or slightly rephrased expressions are acceptable.

### Hallucinated Update

* **factuality error:** The **Generated Memories** includes a new memory related to the "Target Memory for Update," but its content contains factual mistakes or contradictions compared to the correct update.

### Omitted Update

* **Completely omitted:** The **Generated Memories** contains no new memory related to the "Target Memory for Update."
* **Partially omitted:** A related new memory was generated in **Generated Memories**, but it **misses key information** that should have been included.

### Other

Used for update failures that do **not clearly fall** into the above categories of "Hallucination" or "Omission."

# Output Requirements

Please return your evaluation strictly in the following JSON format and provide a concise explanation.

```json
{{
  "reason": "Briefly explain your reasoning here and why it fits this category.",
  "evaluation_result": "Correct | Hallucination | Omission | Other"
}}
```
\end{lstlisting}
\end{tcolorbox}

\begin{tcolorbox}[
  enhanced,
  breakable,
  width=0.98\linewidth,
  colback=tzBlueFill,
  colframe=tzBlueBorder,
  boxrule=1.2pt,
  arc=6pt,
  left=5pt,right=5pt,top=4pt,bottom=2pt,
  title={\small QA Generation Evaluation},
  coltitle=white,
  colbacktitle=tzBlueHeader2,
  fonttitle=\bfseries,
]
\small
\begin{lstlisting}[style=jsonTiny]
You are an **evaluation expert for AI memory system question answering**.
Based **only** on the provided **"Question"**, **"Reference Answer"**, and **"Key Memory Points"** (the essential facts needed to derive the reference answer), strictly evaluate the **accuracy** of the **"Memory System Response."** Classify it as one of **"Correct"**, **"Hallucination"**, or **"Omission."** Do **not** use any external knowledge or subjective inference. Finally, output your judgment **strictly** in the specified JSON format.

# Evaluation Criteria

## Answer Type Classification

### 1. Correct

* The "Memory System Response" accurately answers the "Question," and its content is **semantically equivalent** to the "Reference Answer."
* It contains **no contradictions** with the "Key Memory Points" or "Reference Answer."
* It introduces **no unsupported details** beyond the "Key Memory Points" that could alter the conclusion.
* Synonyms, paraphrasing, and reasonable summarization are acceptable.

### 2. Hallucination

* The "Memory System Response" includes information or facts that **contradict or are inconsistent** with the "Reference Answer" or the "Key Memory Points."
* When the "Reference Answer" is labeled as *unknown/uncertain*, yet the response provides a specific verifiable fact or conclusion.
* Extra irrelevant information that does **not change** the conclusion is **not** considered hallucination by itself; however, if it **changes or misleads** the conclusion, or **contradicts** the "Key Memory Points," it should be judged as a **Hallucination**.

### 3. Omission

* The response is **incomplete** compared to the "Reference Answer."
* It explicitly states "don't know," "can't remember," or "no related memory," even though relevant information exists in the "Key Memory Points."
* For multi-element questions, **all elements must be correct and present**; omission of **any** element is considered an **Omission**.

## Priority Rules (Conflict Handling)

* If the response contains **both missing necessary information** and **fabricated/contradictory information**, classify it as **Hallucination**.
* If there is **no fabrication/contradiction** but some necessary information is missing, classify it as **Omission**.
* Only when the meaning is **fully equivalent** to the reference answer should it be classified as **Correct**.

## Detailed Guidelines and Tolerance

* Equivalent expressions of numbers, times, and units are acceptable, but the **numerical values themselves must not differ**.
* For multi-element questions, **all elements must be complete and accurate**; missing any element counts as **Omission**.
* If the reference answer is *"unknown / cannot be determined"* and the system provides a definite fact, that is a **Hallucination**.
  If the system also answers *"unknown"* (without guessing), it may be **Correct**.
* The evaluation must rely **only** on the *Reference Answer*, *Key Memory Points*, and *System Response* - no external context, world knowledge, or speculative reasoning is allowed.

# Information for Evaluation

* **Question:**
  {question}

* **Reference Answer:**
  {reference_answer}

* **Key Memory Points:**
  {key_memory_points}

* **Memory System Response:**
  {response}

# Output Requirements

Please provide your evaluation result **strictly** in the JSON format below.
Do **not** add any extra explanation or comments outside the JSON block.

```json
{{
  "reasoning": "Provide a concise and traceable evaluation rationale: first compare the system's response with the Key Memory Points (which were correctly used, which were missing, and whether there was any fabrication/contradiction), then assess its consistency with the Reference Answer, and finally state the classification basis.",
  "evaluation_result": "Correct | Hallucination | Omission"
}}
```
\end{lstlisting}
\end{tcolorbox}

\begin{tcolorbox}[
  enhanced,
  breakable,
  width=0.98\linewidth,
  colback=tzBlueFill,
  colframe=tzBlueBorder,
  boxrule=1.2pt,
  arc=6pt,
  left=5pt,right=5pt,top=4pt,bottom=2pt,
  title={\small Retrieval Quality Evaluation},
  coltitle=white,
  colbacktitle=tzBlueHeader2,
  fonttitle=\bfseries,
]
\small
\begin{lstlisting}[style=jsonTiny]
You are a strict **Retrieval Coverage Evaluator**.
Your task is to determine whether a **Retrieved Context** contains enough information to cover a specific **Gold Evidence Point** that is required to answer a question correctly.

# Inputs

1. **Retrieved Context:**
   This is the set of memory entries returned by a retrieval system for a given query.
   {retrieved_context}

2. **Gold Evidence Point:**
   This is the specific key memory fact that *should* be present in the retrieved context in order to answer the question correctly.
   {gold_evidence_point}

# Evaluation Instructions

Determine whether the **Gold Evidence Point** is covered by the **Retrieved Context** using the following scoring rubric:

* **2 - Fully covered:** One or more entries in the retrieved context fully state or logically imply all information in the gold evidence point. Paraphrase and synonymous expressions are acceptable.
* **1 - Partially covered:** Some related information appears in the retrieved context but key details (e.g., a specific value, name, date, or relationship) are missing or imprecise.
* **0 - Not covered:** The retrieved context contains no information about the gold evidence point, or the corresponding content is entirely wrong.

# Scoring Notes

* Focus exclusively on whether the gold evidence point's information is retrievable from the context - do not judge the quality or relevance of other context entries.
* Semantic equivalence is sufficient; exact wording is not required.
* If the retrieved context contains conflicting entries, use the best-matching entry for scoring.

# Output Format

Output **only** the following JSON:

```json
{{
  "reasoning": "Brief explanation of why this score was assigned.",
  "score": "2|1|0"
}}
```
\end{lstlisting}
\end{tcolorbox}

\subsection{Utility Evaluation Prompts}
Utility evaluation is done with LLM-as-a-Judge, where LLMs judge whether the model-generated answer is correct or not by referring to the gold answer. 
\begin{tcolorbox}[
  enhanced,
  breakable,
  width=0.98\linewidth,
  colback=tzBlueFill,
  colframe=tzBlueBorder,
  boxrule=1.2pt,
  arc=6pt,
  left=5pt,right=5pt,top=4pt,bottom=2pt,
  title={\small Answer Evaluation},
  coltitle=white,
  colbacktitle=tzBlueHeader2,
  fonttitle=\bfseries,
]
\small
\begin{lstlisting}[style=jsonTiny]
Your task is to label an answer to a question as 'CORRECT' or 'WRONG'. You will be given the following data:
    (1) a question (posed by one user to another user), 
    (2) a 'gold' (ground truth) answer, 
    (3) a generated answer
which you will score as CORRECT/WRONG.

The point of the question is to ask about something one user should know about the other user based on their prior conversations.
The gold answer will usually be a concise and short answer that includes the referenced topic, for example:
Question: Do you remember what I got the last time I went to Hawaii?
Gold answer: A shell necklace
The generated answer might be much longer, but you should be generous with your grading - as long as it touches on the same topic as the gold answer, it should be counted as CORRECT. 

For time related questions, the gold answer will be a specific date, month, year, etc. The generated answer might be much longer or use relative time references (like "last Tuesday" or "next month"), but you should be generous with your grading - as long as it refers to the same date or time period as the gold answer, it should be counted as CORRECT. Even if the format differs (e.g., "May 7th" vs "7 May"), consider it CORRECT if it's the same date.

Handling "Not answerable" cases:
1. If the GOLD answer is "Not answerable" (meaning the information truly doesn't exist in the conversation history):
   - The generated answer should be CORRECT if it clearly indicates unavailability
   - Accept equivalent expressions: "Not answerable", "There is no information", "There is no direct record", "does not appear to be", "no explicit mention", "cannot be determined", "no specific details available"
   - As long as the generated answer conveys that the information is unavailable, count it as CORRECT

2. If the GOLD answer is a SPECIFIC answer (e.g., "7 May 2023", "John", "Paris"):
   - The generated answer saying "Not answerable" should be counted as WRONG
   - This means the system failed to retrieve information that actually exists in the conversation history
   - Even if phrased as "no information available" or similar, it's still WRONG when the gold answer is specific
   - IMPORTANT: Even if the generated answer mentions the correct information but attributes it to a DIFFERENT person/entity than asked in the question, it should be counted as WRONG. For example, if the question asks about "Alice's opinion" but the answer says "Bob thinks X" (even if X matches the gold answer), this is WRONG because it answers about the wrong person.

3. CRITICAL RULE for "Not answerable" responses:
   - When the generated answer indicates "Not answerable" or similar (cannot find, no information, etc.), the ONLY way it can be CORRECT is if the GOLD answer is ALSO "Not answerable"
   - If the gold answer contains ANY specific information (names, dates, facts, opinions, etc.), then a "Not answerable" response is ALWAYS WRONG, regardless of any explanation or reasoning provided in the generated answer
   - Do NOT be misled by keywords in the explanation - focus on whether the answer actually provides the requested information

Now it's time for the real question:
Question: {question}
Gold answer: {gold_answer}
Generated answer: {generated_answer}

First, provide a short (one sentence) explanation of your reasoning, then finish with CORRECT or WRONG. 
Do NOT include both CORRECT and WRONG in your response, or it will break the evaluation script.

Just return the label CORRECT or WRONG in a json format with the key as "label".
\end{lstlisting}
\end{tcolorbox}

\begin{table*}[t]
    \caption{Additional evaluation of \memguard{} with GPT-4o using the same HaluMem protocol and evaluation settings. Due to computational cost, we evaluate a randomly sampled 20\% subset. $\uparrow$ ($\downarrow$) indicates that higher (lower) values are better. Results marked with $\dagger$ are taken from the original HaluMem evaluation.} 
    \label{tab:halumem_gpt4o}
    \centering
    \scriptsize
    \setlength{\tabcolsep}{3.2pt}
    \resizebox{\textwidth}{!}{
        \begin{tabular}{l*{11}{c}}
            \toprule
            Method &
            \makecell{Extraction\\R $\uparrow$} &
            \makecell{Extraction\\Weighted R $\uparrow$} &
            \makecell{Extraction\\Target P $\uparrow$} &
            \makecell{Extraction\\Acc. (\%) $\uparrow$} &
            \makecell{Extraction\\F1 $\uparrow$} &
            \makecell{Update\\C $\uparrow$} &
            \makecell{Update\\H $\downarrow$} &
            \makecell{Update\\O $\downarrow$} &
            \makecell{Generation\\C $\uparrow$} &
            \makecell{Generation\\H $\downarrow$} &
            \makecell{Generation\\O $\downarrow$} \\
            \midrule
            Mem0$\dagger$
                & 42.91 & 65.03 & 86.26 & 60.86 & 57.31
                & 25.50 & 0.45 & 74.02 & 53.02 & 19.17 & 27.81 \\
            Mem0-Graph$\dagger$
                & 43.28 & 65.52 & 87.20 & 61.86 & 57.85
                & 24.50 & \textbf{0.26} & 75.24 & 54.66 & 19.28 & 26.06 \\
            Zep$\dagger$
                & -- & -- & -- & -- & --
                & 47.28 & 0.42 & 52.31 & 55.47 & 21.92 & 22.62 \\
            MemOS$\dagger$
                & 74.07 & 84.81 & 86.25 & 59.55 & 79.70
                & 62.11 & 0.42 & 37.48
                & \textbf{67.23} & \textbf{15.17} & \textbf{17.59} \\
            Memobase$\dagger$
                & 14.55 & 25.88 & 92.24 & 32.29 & 25.13
                & 5.20 & 0.55 & 94.25 & 35.33 & 29.97 & 34.71 \\
            Supermemory$\dagger$
                & 41.53 & 64.76 & 90.32 & 60.83 & 56.90
                & 16.37 & 1.15 & 82.47 & 54.07 & 22.24 & 23.69 \\
            \midrule
            \memguard{} (GPT-4.1)
                & \textbf{90.47} & \textbf{93.95} & \textbf{98.13}
                & \textbf{89.53} & \textbf{94.15}
                & \textbf{70.79} & 0.35 & \textbf{28.86}
                & 59.50 & 16.32 & 24.17 \\
            \memguard{} (GPT-4o, 20\%)
                & 82.44 & 89.93 & 87.23 & 74.53 & 84.76
                & 66.40 & 0.67 & 52.93
                & 63.37 & 16.93 & 19.70 \\
            \bottomrule
        \end{tabular}
    }
\end{table*}

\section{Additional Experimental Results}
\subsection{Additional GPT-4o Evaluation on HaluMem.}
\label{app:additional_exp_halumem}
\paragraph{Additional GPT-4o Evaluation on HaluMem.}
To assess sensitivity to the base LLM, we additionally evaluate \memguard{} with GPT-4o using the same HaluMem protocol and evaluation settings. Due to computational cost, this evaluation is conducted on a randomly sampled 20\% subset. As shown in \Cref{tab:halumem_gpt4o}, \memguard{} achieves 74.53 extraction accuracy and 84.76 extraction F1, exceeding the strongest reported baselines by 12.67 and 5.06 points, respectively. It also obtains 66.40 update correctness and 63.37 generation correctness, with generation hallucination and omission rates of 16.93 and 19.70. Although some metrics differ from the GPT-4.1 results, the consistently strong performance across extraction, update, and generation indicates that \memguard{}'s gains are not specific to a single base LLM, supporting the robustness of its unified contamination-control pipeline.

\subsection{Memory-Type Classification Reliability.}
Because memory-type assignments determine both type-isolated storage and query-adaptive retrieval, classification errors can directly weaken \memguard{}'s functional boundaries. We therefore randomly sample 100 atomic memory entries, manually annotate their functional roles, and evaluate the same base LLM used in our experiments (\texttt{GPT-4.1-mini}) against these human labels. The classifier achieves 89\% accuracy, indicating that semantic, episodic, and procedural memories can be distinguished reliably in practice. We do not claim that this three-type taxonomy is universally sufficient for every memory system; rather, we adopt it based on three complementary forms of evidence: established theoretical distinctions among the three functional memory roles, empirical type-confusion analysis showing that these boundaries capture prevalent contamination mechanisms, and classifier evaluation classifier evaluation demonstrating that the taxonomy can be operationalized with high agreement.

\section{Use of Large Language Models}
\label{Appendix:llm_usage_statement}
Large language models, such as ChatGPT, are used exclusively for grammar checking during the writing process. They are not used for research ideation.
\end{document}